\documentclass[lettersize,journal]{IEEEtran}
\usepackage{amsmath,amsfonts}
\usepackage{algorithmic}
\usepackage{algorithm}
\usepackage{array}
\usepackage[caption=false,font=normalsize,labelfont=sf,textfont=sf]{subfig}
\usepackage{textcomp}
\usepackage{stfloats}
\usepackage{url}
\usepackage{verbatim}
\usepackage{graphicx}
\usepackage{cite}
\usepackage{pdfpages}

\def\hideResults{0} 
\def\draftFigures{0} 

\usepackage{setspace}
\usepackage[switch]{lineno}
\usepackage{comment}
\usepackage{amsmath}
\usepackage{amssymb}
\if\draftFigures1
  \usepackage[draft]{graphicx}
\else
  \usepackage{graphicx}
\fi
\graphicspath{{Figures/}}
\usepackage{hyperref}
\usepackage{float}
\usepackage{multirow}
\usepackage{afterpage}
\usepackage{etoolbox}
\usepackage{hhline,colortbl}
\usepackage{tablefootnote}
\usepackage{caption}

\definecolor{todoColor}{rgb}{0.0,0.0,1.0}
\newcommand{\todo}[1]{\color{todoColor}[TODO: {#1}]\normalcolor}

\definecolor{meetingColor}{rgb}{0.0,0.7,0.0}
\newcommand{\meeting}[1]{\color{meetingColor}[From meeting notes: {#1}]\normalcolor}

\definecolor{noteToSelfColor}{rgb}{0.6,0.6,0.6}
\newcommand{\noteToSelf}[1]{\color{noteToSelfColor}{#1}\normalcolor}
\iftrue
  \renewcommand{\noteToSelf}[1]{}
\fi

\definecolor{todoDoneColor}{rgb}{0.0,0.6,0.0}

\definecolor{unresolvedColor}{rgb}{0.6,0.0,0.0}

\definecolor{questionColor}{rgb}{1.0,1.0,0.0}

\definecolor{answerColor}{rgb}{0.0,1.0,0.0}

\newcommand{\fat}[1]{\mathbf{#1}} 
\newcommand{\bldgr}[1]{\boldsymbol{#1}} 
\newcommand{\transp}{^T} 

\newcommand{\argmax}{\operatornamewithlimits{argmax}}
\newcommand{\argmin}{\operatornamewithlimits{argmin}}

\newlength{\mywidth}
\newlength{\myheight}
\newlength{\myspace}

\begin{document}

\title{An Out-Of-The-Box Mutual Information Nonlinear Registration and Uncertainty Quantification method}

\title{Efficient Zero-shot Probabilistic Image Registration Using Latent Variables}
\title{Efficient Multimodal Image Registration with Error Bars Using Latent Variables}
\title{A Latent-Variable Model for Efficient Adaptive Registration with Uncertainty}
\title{A Latent-Variable Model for Generalizable Registration with Efficient Uncertainty Quantification}
\title{A Latent-Variable Model for Generalizable Multimodal Registration with Uncertainty Quantification}
\title{A Latent-Variable Model for Out-of-the-Box Multimodal Registration with Uncertainty Quantification}
\title{Out-of-the-Box Multimodal Registration with Uncertainty Quantification Using a Latent-Variable Model}
\title{BINDER: 
A Latent Variable Model for Probabilistic Medical Image Registration}


\author{Stefano Cerri*, 
Amirhossein Hassankhani*, 
Ya\"{e}l Balbastre, Koen Van Leemput
\thanks{* Joint first authorship}
\thanks{Stefano Cerri is with the Copenhagen Research Centre for Biological and Precision Psychiatry, Mental Health Centre Copenhagen, Copenhagen University Hospital, Denmark, and the Pioneer Centre for Artificial Intelligence, University of Copenhagen, Denmark (e-mail: stce@di.ku.dk).}
\thanks{Amirhossein Hassankhani is with the Department of Neuroscience and Biomedical Engineering, Aalto University, Finland.}
\thanks{Ya\"{e}l Balbastre is with the Department of Experimental Psychology, Division of Psychology and Language Sciences, University College London.}
\thanks{Koen Van Leemput is with the Departments of Neuroscience and Biomedical Engineering and of Computer Science at Aalto University, Finland.}
}

\maketitle
\bstctlcite{IEEEexample:BSTcontrol}

\begin{abstract}


We propose 
a new probabilistic model for 
general-purpose
medical
image registration that 
builds upon
the 
mutual information registration criterion.
%
%
It 
centers around
a spatial interpolation 
technique
that assumes
latent voxel-wise correspondences between the images being registered. 
By exploiting these latent variables,
we 
derive 
dedicated optimization and 
MCMC sampling
techniques 
that 
only involve closed-form iterative updates.
%
%
%
When applied to nonlinear registration, 
an 
efficient 
demons-like
optimization algorithm
is obtained 
that 
shows
robust 
out-of-the-box
performance
across a variety of 
monomodal and multimodal
registration
tasks. 
%
We also 
demonstrate 
a 
corresponding
sampler 
that can
quantify,
for the first time,
uncertainty
in
multimodal registration
scenarios 
with
very high-dimensional 
3D deformations.
%

Our code, which we call BINDER (Bayesian INference for DEformable Registration), is freely available at \url{https://github.com/ste93ste/BINDER}.

\end{abstract}


\section{Introduction}

Medical image registration -- the process of aligning and deforming two images to bring their anatomical content into correspondence -- is a core technique in medical image analysis, with numerous applications in research and clinical practice.
%
In routine clinical applications, where image characteristics such as the tissue contrast, the field of view and the voxel size vary unpredictably due to 
changes in imaging 
equipment and acquisition protocols,
the \emph{generalization} ability of 
automatic
registration algorithms 
is 
of vital importance.
%
Registration results
should ideally 
also
come with 
automatic
measures of \emph{uncertainty} (error bars),
as 
this
may
facilitate clinical acceptance,
enable automatic quality control,
and allow registration uncertainty to be propagated to downstream applications such as surgery or radiotherapy~\cite{Bierbrier2022}.

Despite decades of research 
(see~\cite{Sotiras2013,Viergever2016,
Chen2025} for detailed reviews of the existing literature), 
fundamental
issues pertaining to the \emph{generalization} and \emph{uncertainty quantification} abilities 
of medical image registration techniques
persist.

\noteToSelf{ 

Bierbrier2022: "The product of a registration is useful for numerous downstream clinical and research tasks including motion correction, motion de- termination, cross modality image fusion, change detection, distor- tion correction, atlas construction, atlas registration and segmentation (Holden, 2008)."

Bierbrier2022: "the variable performance of deformable image registration impedes its clinical acceptance"

Bierbrier2022: ``we review methods that densely estimate the deformable image registration error, or confidence, between two registered images, without any manual intervention.''

Bierbrier2022: "The goal is to be able to automatically and quantitatively assess the results of any given registration immediately after it is performed. This could enable, for example, a surgeon to place trust in a region of a registration with low estimated error (or confidence) during an image-guided [...] or radiation therapy [...] procedure. 
[...] It can vastly decrease manual assessment time and enable automatic quality control of large-scale image analysis [...]"

TMI special issue: ``[methods for \ldots] often suffer from limited generalization, failing to maintain performance when applied to imaging modalities or sampling patterns that differ from their training data''

Domingos: ``The \#1 problem in AI is poor generalization''

}

\subsection{Limits to generalization}

%
Over the last few decades, a number of registration metrics with excellent generalization properties have 
emerged,
including those based on cross-correlation~\cite{Avants2008}, mutual information (MI)~\cite{Wells1996,Maes1997}, and other principles~\cite{Heinrich2012}. Such metrics are able to quantify how well two images align without 
putting any constraints on the 
characteristics of those images, such as their respective image size, voxel resolution, contrast properties or even the organs that 
they depict. 
This allows the registration problem to be re-formulated as an optimization problem that can be solved 
for 
any image pair, 
enabling robust performance across a wide gamut of applications 
-- in theory
without making any changes to the 
registration software 
that is used.
%
%

In practice, however, the user 
often
still
needs to 
adjust several configuration settings in order to attain the best results 
for any particular 
application. 
Such settings may include
the optimization algorithm 
that is used
and how it is tuned, 
the 
way spatial transformations are parameterized,
various regularization strengths,
and even the registration metric 
itself.
In academic benchmarks, 
expert-provided
annotations 
-- 
such as segmentation masks or corresponding landmarks 
--
are typically 
used to adjust
each algorithm's
settings on 
held-out
images 
that are well-matched with 
the images that are used to compute the rankings.
Given the difficulty in obtaining such annotations in 
clinical scenarios, 
however,
this practice may obscure some of the limitations on generalization 
of
these methods, and inflate expectations of their out-of-the-box real-world performance. 



Instead of optimizing a registration metric for each pair of images independently, 
a
global
mapping between image pairs and 
spatial transformations 
that optimize
the
metric 
can also be learned 
by a neural network~\cite{Balakrishnan2019,Devos2019}. 
%
Once trained, this procedure 
can substantially
speed up the registration process, because the learned mapping can simply be applied to 
a
new image pair
in a single forward operation.
However, it also leads to a 
loss of generalizability:
accuracy
typically
degrades 
when 
images with characteristics 
different from those 
in the training dataset
need to be
registered,
even 
when
the 
original 
registration 
metric 
itself
generalizes well~\cite{Hoffmann2022,Jena2024,Chen2025}.
%
When they are available,
expert annotations 
can also be exploited during network training,
by optimizing 
for the alignment of clinically meaningful structures 
instead of traditional registration metrics~\cite{Hu2018}.
While this 
often improves
registration accuracy, 
it also 
limits the practicality of adaptation
to new anatomies or imaging domains, 
as new annotations for retraining may then be required~\cite{Jena2024}.


The 
limited 
generalization abilities of learning-based registration can be substantially 
improved
by using diverse synthetic data in the training procedure to expose the networks to a broader range of shapes and contrasts than would be found in a real dataset~\cite{Hoffmann2022}.
%
Nevertheless, these approaches remain amortized over a training set. The resulting models therefore encode statistical assumptions about the training distribution, and do not recover the exact behavior of optimization-based approaches. Importantly, errors in learning-based registration have been shown to be mostly driven by the statistical mismatch between training and testing sets~\cite{ketcha2019learning}.

\noteToSelf{

Balakrishnan2019: ``Essentially, we replace pair-specific optimization of the deformation field by global optimization of the shared parameters, which in other domains has been referred to as amortization [12]–[15].''

Balakrishnan2019: ``Our method substitutes the pair-specific optimization over
the deformation field [\ldots] with a global optimization of function
parameters [\ldots] for function [\ldots]. This process is sometimes
referred to as amortized optimization [66].'' 

}

\subsection{Difficulties in quantifying uncertainty}

While other 
techniques
exist (e.g.,~\cite{%
HubTMI2009,
HubPMB2013,
LoftiMLMI2013,
HeinrichMEDIA2016,
YangNI2017,
SedghiUNSURE2019}), 
the task of 
characterizing
registration
uncertainty is typically 
addressed
within the context of 
generative
probabilistic 
models~\cite{%
KybicISBI2008,
GeeIPMI1995,
RisholmWBIR2010,
RisholmMEDIA2013,
PursleyMP2012,
RisholmMICCAI2011,
SimpsonNI2012,
SimpsonMEDIA2015,
LeFolgocMEDIA2017,
LeFolgocTMI2017,
YangMICCAI2015,
DalcaMEDIA2019,
SchultzSPIE2018,
SchultzSPIE2019,
WangMICCAI2018,
WatanabeWBIR2012,
Grzech2021}.
In this approach, 
a prior distribution over spatial transformations 
(encoding 
domain knowledge
such as their expected spatial smoothness)
is combined with a 
%
%
likelihood function 
that models
how one image is generated 
from another 
given
a 
particular
spatial transformation.
%
The aim is then to apply Bayes' rule to characterize the entire posterior distribution 
of the spatial transformation 
between 
a pair of
images 
-- not just its mode (the most probable 
registration outcome%
)
but also all plausible alternative solutions.

%
The current state of the art in probabilistic image registration suffers from two 
major
limitations. 
First,
most
methods
assume voxel-wise Gaussian noise (corresponding to a sum-of-squared-differences registration metric)~\cite{%
RisholmWBIR2010,
RisholmMEDIA2013,
RisholmMICCAI2011,
SimpsonNI2012,
SimpsonMEDIA2015,
YangMICCAI2015,
DalcaMEDIA2019,
WangMICCAI2018} 
or a 
variant thereof~\cite{%
KybicISBI2008,GeeSPIE1995,
LeFolgocMEDIA2017,
LeFolgocTMI2017,
Grzech2021} 
for the likelihood function, which 
restricts their applicability to monomodal registration 
settings 
only,
excluding a large number of applications in which multimodal registration 
is required.
%
Second,
dealing with posterior distributions over 
dense
3D deformation fields 
has
remained 
%
challenging
due to the extremely high dimensionality of the problem.
%
An
often-used
approach
is to approximate the posterior with a 
multivariate Gaussian distribution,
in the simplest case by matching the local curvature of the log-posterior at the most probable solution (so-called Laplace approximation)~\cite{%
KybicISBI2008,
YangMICCAI2015,
WangMICCAI2018,
WatanabeWBIR2012}.
%
In a more advanced version
a variational approach is taken in which
the 
Gaussian
mean and variance 
are 
determined by numerically minimizing the Kullback-Leibler (KL) divergence between the 
approximation and the full posterior~\cite{%
SimpsonNI2012,
SimpsonMEDIA2015,
LeFolgocMEDIA2017,
DalcaMEDIA2019,
SchultzSPIE2019}.
%
However, Gaussian approximations of the 
posterior
cannot capture such 
aspects
as multiple modes, heavy tails, and other non-Gaussian characteristics
that may occur in real-world registration applications (see Figs.~\ref{fig:case4} and~\ref{fig:case2} in this paper for examples).
In such cases
variational approximations
can severely 
under-represent
registration
uncertainty,
since
putting probability mass in
regions with low posterior probability 
is
heavily penalized by the KL divergence~\cite{Bishop2006chapter10} 
--
see Fig.~9 in~\cite{LeFolgocTMI2017} for an example of this effect.
This 
becomes especially 
pronounced
when the variance is 
further
assumed to be diagonal~\cite{DalcaMEDIA2019}
or otherwise heavily restricted~\cite{%
KybicISBI2008,
YangMICCAI2015,
WatanabeWBIR2012}.
But 
relaxing
such
restrictions
is only feasible for 
low-dimensional deformation models,
such as 
those using
a
modest
number of 
basis functions~\cite{%
SimpsonNI2012,
SimpsonMEDIA2015,
LeFolgocMEDIA2017,
WangMICCAI2018},
because of the computational 
difficulty
of storing and 
decomposing the resulting full precision matrices.

The representational limitations of Gaussian approximations can be avoided by resorting to 
Markov chain Monte Carlo (MCMC) 
sampling methods~\cite{%
GeeIPMI1995,
RisholmWBIR2010,
RisholmMEDIA2013,
PursleyMP2012,
RisholmMICCAI2011,
LeFolgocTMI2017,
Grzech2021,
SchultzSPIE2018}, 
which
can in principle generate unbiased samples from the true posterior  distribution
of arbitrarily complex deformation fields. 
In practice, however, they 
too 
have typically been applied only to registration models with a limited number of 
degrees of freedom.
This is because 
designing effective 
Metropolis-Hastings
proposal distributions 
in high dimensions
is 
very
challenging:
updating 
all
parameters simultaneously 
(e.g., using random draws from a multivariate Gaussian with non-diagonal variance~\cite{%
RisholmMEDIA2013}) 
can become
computationally expensive and difficult to tune
when there are many parameters,
whereas updating only one parameter at a time~\cite{%
GeeIPMI1995,
LeFolgocTMI2017} can 
become
very
inefficient.

\noteToSelf{

Kingma: ``KL divergence of the approximate from the true posterior''

Bishop: ``Kullback-Leibler divergence between [approx] and the posterior distribution''

Bishop: ``It is a general result that a factorized variational approximation tends to give approximations to the posterior distribution that are too compact.''

Bishop: ``The difference between these two results can be understood by noting that there is a large positive contribution to the Kullback-Leibler divergence [...] from regions of Z space in which p(Z) is near zero unless q(Z) is also close to zero. Thus minimizing this form of KL divergence leads to distributions q(Z) that avoid regions in which p(Z) is small.''

Bishop: ``In practical applications, the true posterior distribution will often be multimodal, with most of the posterior mass concentrated in some number of relatively small regions of parameter space. [...] 

Bishop: ``[...] a variational treatment based on the minimization of KL(q|p) will tend to find one of these modes.''

Wikipedia: ``[careful: KL(p|q) so the expection-propagation one!] a measure of how much an approximating probability distribution Q is different from a true probability distribution P''

Bishop: ``The most serious limitation of the Laplace framework, however, is that it is based purely on the aspects of the true distribution at a specific value of the variable, and so can fail to capture important global properties. In Chapter 10 we shall consider alternative approaches which adopt a more global perspective.''

}

\subsection{Contributions}


The aim of this paper is to help advance the 
state of the art in registration \emph{generalizability} (the ability to work directly out-of-the-box on any pair of images) while simultaneously enabling \emph{uncertainty quantification} (the ability to characterize a lack of information about the ``correct'' registration solution) in such settings.
Towards this end, we make the following contributions:

\begin{enumerate}

  \item \emph{Probabilistic registration in multimodal settings:}
  We expand the scope of 
  probabilistic registration methods,
  by replacing the 
  Gaussian noise model that currently confines these methods to monomodal settings
  with a probabilistic reformulation of the 
  MI
  registration criterion. 
  Importantly, this reformulation involves
  a new spatial interpolation model that 
  assumes
  latent voxel correspondences between the images being registered. 
  Conditioning on those latent correspondences 
  decomposes the posterior of the model parameters into easy-to-handle individual factors, 
  enabling 
  both effective
  parameter optimization and uncertainty quantification
  in general multimodal settings.
  

  \item \emph{Robust out-of-the-box generalization:}
  For conventional optimization-based registration,
  the latent correspondences in the model can be exploited to devise an efficient expectation-maximization (EM) optimization algorithm. Since 
  it
  only involves iterative updates that are given in analytical form, 
  any requirement for tuning the optimizer by the user is avoided.
  We show experimentally that, 
  when this technique is applied to nonlinear registration with curvature-based regularization, competitive registration accuracies 
  can be
  obtained that are more consistent across organs, modalities and acquisition protocols than those of several established benchmark methods. 
  Furthermore, sensitivity to the exact value of the 
  one remaining hyperparameter of the method (the strength of the regularization)
  appears to be lower than the sensitivity of the benchmark methods to their respective hyperparameter settings. 
  
  
  \item \emph{
  Sampling of 
  free-form
  3D deformation fields:}
  The convenient factorization properties of the posterior 
  also enable a new MCMC method for uncertainty quantification, in which some variables are directly updated by conditioning on the remaining variables and vice versa (Gibbs sampling).
  This makes the sampler work out-of-the-box without any form of tuning,
  and avoids approximating the distribution over deformations with a functional (e.g., Gaussian) form.
  Furthermore, 
  we demonstrate experimentally that it becomes feasible to sample from dense 3D deformation field models with millions of parameters -- 
  orders of magnitude higher than the models 
  typically
  reported 
  in the literature
  \footnote{The approximate Langevin Monte Carlo-inspired sampler of~\cite{Grzech2021} also addresses very high-dimensional deformation fields, but 
  seems to underestimate uncertainty even compared to 
  a heavily-restricted variational 
  approach. 
  }.
  
    %

\end{enumerate}

An early version of this work,
introducing
the conceptual framework,
was presented 
in~\cite{Agn2019}. 
Here we significantly expand on that paper, with detailed derivations, an 
efficient
GPU implementation that we make freely available,
and a large number of experiments and benchmark comparisons in different organs, image modalities and acquisition parameters.

\subsection{Related work}

Several specific connections of the proposed approach with prior art are worth 
mentioning.

\emph{Demons algorithm:}
When used for nonlinear registration,
the proposed EM optimizer
can be seen as 
an alternative to
the 
well-known
demons algorithm~\cite{Thirion1998},
in which latent 
spatial
correspondences are repeatedly computed and 
regularized using 
fast 
filtering techniques~\cite{Cachier2003}. 
%
But whereas the demons algorithm 
relies on a linearization inside the sum-of-squared-differences (SSD) registration criterion 
to obtain its correspondences
--
and can therefore only be used
for monomodal registration
--
the proposed 
algorithm
provides
a similar
efficient optimization strategy 
for general multimodal registration.

\emph{Point set registration:}
Our EM optimization
is based on 
an
iterative 
statistical
association
of the 
voxels
in one image 
with
those in the other image, which then drives an update of the transformation parameters 
and vice versa.
%
As such,
this approach can be interpreted as a 
registration of two point clouds in the style of the iterative closest point algorithm~\cite{Besl1992}, in particular when it is solved using EM optimization~\cite{Granger2002,Chui2000}. 

\emph{Partial volume interpolation:}
In one of the original formulations of the MI registration criterion~\cite{Maes1997}, 
a 
trilinear
``partial volume distribution'' interpolation strategy was introduced 
to ensure that
the registration criterion 
changes smoothly
as
the transformation parameters are varied.
In this approach, 
and in later generalizations~\cite{Chen2003},
the contribution of 
one voxel
is distributed 
over 
several 
histogram bins
according to a 
B-spline interpolation kernel.
%
The same type of 
distribution
takes place in our 
method,
but
with distribution weights that 
also
take
intensity (in)compatibility
into account.

\section{Generative Model}


To simply notation, we only describe the proposed methods in 1D here.
We refer the reader to Sec.~VII of the supplementary material for the extension to higher dimensions.

%
Let $\fat{u} = (u_1, \dots, u_I)^T$ be a ``fixed'' image with $I$ voxels, where the intensities $u_i \in \{ 1, \dots, L \}$ can take $L$ discrete values. We 
will
model $\fat{u}$ as being generated 
by deforming
a ``moving'' image $\fat{v} = (v_1, \dots, v_J)^T$ with $J$ voxels, with intensities $v_i \in \{ 1, \dots, K \}$ taken from $K$ discrete levels.
For this purpose,
we 
map the initial spatial position $x_i$ of voxel $i$ 
(measured in voxel coordinates in the image grid of $\fat{v}$)
to a new position 
\begin{equation*}
y_i 
= x_i + \bldgr{\phi}_i\transp \fat{c},
\end{equation*}
where $\bldgr{\phi}_i = (\phi_{i,1}, \ldots, \phi_{i,M})\transp$ evaluates $M$ basis functions in voxel $i$, and $\fat{c} \in \mathbb{R}^M$ are spatial transformation parameters. 

Below
we first sketch a simple 
conceptual 
framework
that has a direct link with the MI registration criterion.
This 
framework,
which is due to~\cite{Roche2000}, is then extended 
to a 
full
Bayesian model
for 
out-of-the-box
multimodal image
registration.

For reasons that will 
soon 
become clear, in the remainder we 
refer to the voxels in the moving image and to their intensities as ``nodes'' and ``classes'', respectively.
Since the moving image is a constant in the generative process,
explicit dependencies on $\fat{v}$ are omitted from all equations for notational convenience.

\subsection{Model sketch: Mutual Information}


In the ``unspecified channel'' model of~\cite{Roche2000},
%
%
each class $k$ is associated with a categorical (discrete) distribution 
$\mathrm{Cat}(\bldgr{\theta}_k)$ over the $L$ intensity levels,
parameterized by $\bldgr{\theta}_k = (\theta_{k,1}, \ldots, \theta_{k,L})$ with $\theta_{k,l}$ the probability of intensity $l$.
We collect the parameters of these distributions in the variable $\bldgr{\theta} = \{\bldgr{\theta}_k\}_{k=1}^K$.
For given transformation parameter values $\fat{c}$,
each voxel $i$ in $\fat{u}$ is 
assigned
to
the node in $\fat{v}$ that is spatially nearest to its mapped location $y_i$.
The class of that node is then recorded, 
and 
an intensity $u_i$ is drawn randomly from the 
corresponding categorical distribution. 
%
Note that the moving image is 
conceptually
padded to infinite size
in this procedure, with virtual nodes outside its original field-of-view having a constant intensity~\cite{Wells1996,Roche2000} (assumed $k=1$ in the remainder of this paper).

%
Counting the number of voxels with intensity $l$ 
assigned
to nodes of class $k$ 
as
$N_{k,l}(\fat{c})$,
and dropping the dependency on $\fat{c}$ to declutter notation,
the log-likelihood 
of the model parameters $\{\fat{c}, \bldgr{\theta}\}$
can be written as:
$$
\log p( \fat{u} | \fat{c}, \bldgr{\theta} ) = \sum_{k,l} N_{k,l} \log \theta_{k,l}
.
$$
Maximizing this log-likelihood 
is 
equivalent to registering the fixed and moving image using the MI registration criterion, because
%
%
%
\begin{eqnarray*}
\max_{\fat{c},\bldgr{\theta}} 
\left[
\log p( \fat{u} | \fat{c}, \bldgr{\theta} ) 
\right]
& = &
\max_{\fat{c}} 
\left[
\max_{\bldgr{\theta}} 
\left[
\log 
p(\fat{u} | \bldgr{\theta}, \fat{c})
\right]
\right] 
\\
& = &
\max_{\fat{c}} 
\left[
\sum_{k,l}
N_{k,l} \log 
\left(
\frac{N_{k,l}}
{\sum_l N_{k,l}}
\right)
\right]
\\
& = &
\max_{\fat{c}} 
\left[
\sum_{k,l}
n_{k,l} \log \frac{n_{k,l}}{n_k n_l} 
\right]%
,
\end{eqnarray*}
where 
$n_{k,l} = N_{k,l}/I$,
$n_k = \sum_l n_{k,l}$
and $n_l = \sum_k n_{k,l}$
are normalized counts in the joint and marginal histograms, respectively.
(In the last step 
the marginal entropy 
$
-\sum_l n_l \log n_l%
$
was introduced, which does not depend on $\fat{c}$.)


%

\subsection{Full Model}


Although conceptually appealing, the maximum likelihood formulation is not yet sufficient for practical registration with uncertainty quantification: 
its nearest-neighbor interpolation scheme introduces abrupt changes in the registration criterion as the transformation parameters are varied,
impeding numerical optimization and limiting registration accuracy,
and prior distributions on the model parameters are missing.

\begin{figure*}[!ht]
  \setlength{\mywidth}{.32\linewidth}
  \centering
  \begin{tabular}{ccc}
    \raisebox{-.5cm}{\includegraphics[width=\mywidth]{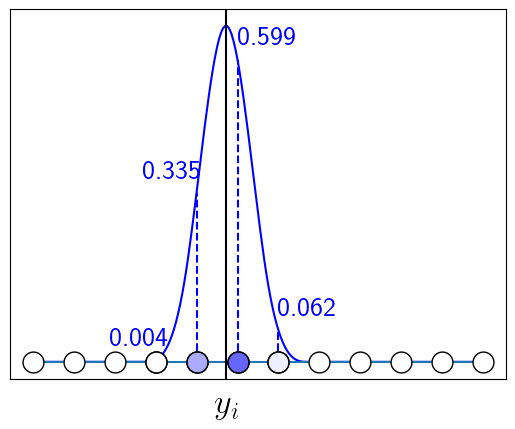}} &
    \includegraphics[width=\mywidth]{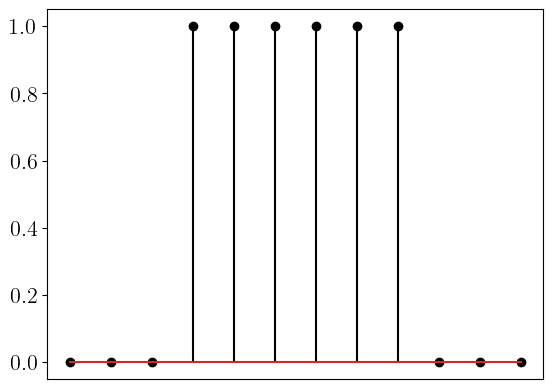} &
    \includegraphics[width=\mywidth]{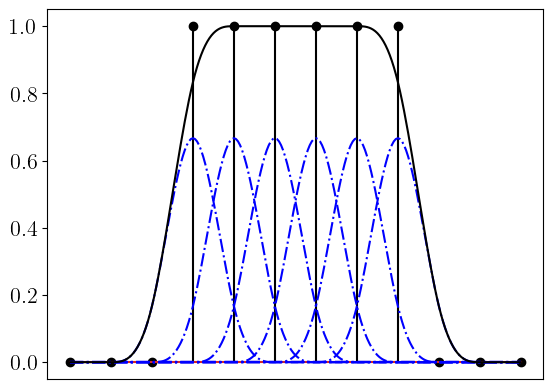}
  \end{tabular}
  \captionsetup{singlelinecheck=off} 
  \caption[.]{Illustration of the spatial interpolation model in 1D. A voxel $i$ is 
  randomly
  assigned to a node $j$ with probability $\beta(j-y_i)$, so that nearby nodes have a higher probability of being selected than nodes farther away (left). 
  Marginalizing over the node assignments turns a discrete one-hot encoding for a class $k$ (middle) into a continuous probability map $\pi_k(y)$ (right).}
  \label{fig:Bspline_interp}
\end{figure*}

To address the interpolation issue, 
we introduce a latent variable $\fat{n} = (n_1, \ldots, n_I)\transp$ 
in the model,
where $n_i \in \{1, \ldots, J\}$ denotes the node that voxel $i$ is assigned to.
%
Then, instead of deterministically assigning each voxel $i$ to the node 
closest to its currently mapped location $y_i$, 
we 
center a standard cubic B-spline $\beta(\cdot)$ around $y_i$ and assign 
it
stochastically,
with probability
$$
p( n_i = j | \fat{c} ) = \beta( j - y_i)
$$
of node $j$ being selected.
This process is illustrated in Fig.~\ref{fig:Bspline_interp} (left); 
it is a valid probabilistic model because $\sum_{j \in \mathbb{Z}} \beta( j - y ) = 1, \forall y$.
%
Note that in practice only a limited number of nodes (4 in 1D, 64 in 3D) have a non-zero probability of being assigned to $i$ because B-splines have limited 
spatial
support.

%
Since 
$$
p( \fat{u} | \fat{n}, \bldgr{\theta} ) = \prod_i p( u_i | n_i, \bldgr{\theta} ) 
\hspace{1ex}
\text{with}
\hspace{1ex}
p( u_i | n_i = j, \bldgr{\theta} ) = \theta_{v_j,u_i}
$$
and 
$$
p( \fat{n} | \fat{c} ) = \prod_i p( n_i | \fat{c} ),
$$
we can marginalize analytically over $\fat{n}$ to obtain the following likelihood function:
\begin{eqnarray*}
p( \fat{u} | \fat{c}, \bldgr{\theta} ) 
& = & \sum_{\fat{n}} p( \fat{u} | \fat{n}, \bldgr{\theta} ) p( \fat{n} | \fat{c} )
\\
& = &
\prod_i 
\left( 
\sum_k \pi_k( y_i )
\theta_{k,u_i} 
\right)
,
\end{eqnarray*}
where
\begin{align*}
  \pi_k( y ) = \sum_j [ v_j = k ] \beta( y - j ), \quad \forall k 
  .
\end{align*}
Here $[\cdot]$ denotes the Iverson bracket, which evaluates to $1$ if its argument is true, and to $0$ otherwise.
The effect of the stochastic node assignment procedure can therefore be interpreted as 
a spatial
interpolation model in which $K$ discrete, one-hot encoding images are extracted from the 
original moving image, and then converted into $K$ continuous spatial maps.
%
%
Each spatial map $\pi_k(y)$ 
represents the probability with which class $k$ occurs at 
continuous
locations $y$, as illustrated in Fig.~\ref{fig:Bspline_interp} (right) and Fig.~7 of the supplementary material.
The resulting generative 
process
is therefore similar to the one used in Bayesian segmentation models using deformable probabilistic atlases:
the class in each voxel is drawn randomly from deformed probability maps, and an intensity is generated accordingly~\cite{Ashburner2005,Puonti2016}.

%
%
%
%

To obtain a complete Bayesian model,
we 
also
need to define prior distributions over the model parameters.
For $\bldgr{\theta}$ we use a conjugate prior of the form
$
p( \bldgr{\theta} ) = \prod_k p( \bldgr{\theta}_k )
$
with
$$
p( \bldgr{\theta}_k )
= 
\mathrm{Dir}( \bldgr{\theta}_k | N^0_{k,1}+1, \ldots, N^0_{k,L}+1).
$$
Here
$
\mathrm{Dir}( \bldgr{\theta}_k | \alpha_1, \ldots, \alpha_L ) \propto \prod_l \theta_{k,l}^{\alpha_{l}-1}
$
denotes a Dirichlet distribution, 
and 
$N^0_{k,l}$ 
can be interpreted as ``prefills'' of the bins in a joint histogram,
set to
$N^0_{k,l} = 1, \forall k,l$ in this paper.
For 
$\fat{c}$
we use a Gaussian prior of the form
\begin{equation}
p( \fat{c} ) = \mathcal{N}(\fat{c} | \fat{0}, (\gamma \fat{P})^{-1} ),
\label{eq:c_prior}
\end{equation}
where $\fat{P}$ is 
a
precision matrix 
that aims to penalize undesirable values of $\fat{c}$,
and $\gamma$
is a tunable hyperparameter that determines the strength of the penalty.

\section{Inference}


Given a pair of images $\fat{u}$ and $\fat{v}$, 
the 
values of the
model parameters $\{\fat{c},\bldgr{\theta}\}$ 
can be inferred from
their joint posterior distribution
$
p( \fat{c}, \bldgr{\theta} | \fat{u} )
\propto
p( \fat{u} | \fat{c}, \bldgr{\theta} )
p( \fat{c} ) p( \bldgr{\theta} )
,
$
using either optimization to obtain point estimates 
as in MI-based registration,
or MCMC sampling to quantify registration uncertainty.

\begin{figure}[!t]
  \centering
  \includegraphics[width=.75\linewidth]{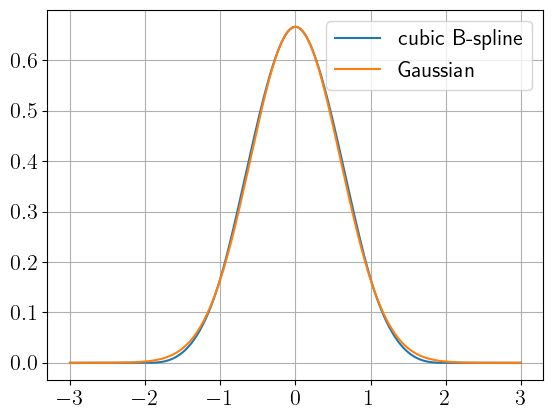}
  \captionsetup{singlelinecheck=off} 
  \caption[.]{A cubic B-spline 
  $\beta(z)$
  is well approximated by a Gaussian 
  $\mathcal{N}( z | 0, \sigma^2 )$
  with variance
  $\sigma^2 = 9/(8 \pi)$.
  The variance is chosen so that $\beta(0) = \mathcal{N}( 0 | 0, \sigma^2 )$.}
  \label{fig:Gaussian_approx}
\end{figure}

Although the node assignments $\fat{n}$ are 
marginalized out 
in the likelihood function, 
exploiting them
explicitly 
dramatically simplifies
inference.
This is because their 
values 
are readily
inferred 
for a given set of model parameters:
\begin{equation*}
p(\fat{n} | \fat{u}, \bldgr{\theta}, \fat{c} )
= 
\prod_i p( n_i | u_i, \bldgr{\theta}, \fat{c} )
\end{equation*}
with
\begin{equation}
p( n_i=j | u_i, \bldgr{\theta}, \fat{c} ) 
\, \propto \,
\theta_{v_j,u_i} \beta( y_i - j ),
\end{equation}
which involves evaluating only a few candidate nodes for each voxel. 
%
In turn,
conditioning on
$\fat{n}$
completely
decouples the 
model parameters from each other:
\begin{equation*}
p( \fat{c}, \bldgr{\theta} | \fat{u}, \fat{n} ) 
= 
p( \fat{c} | \fat{n} )
\prod_k p( \bldgr{\theta}_k | \fat{u}, \fat{n} )
.
\end{equation*}
Here
\begin{equation*}
p( \bldgr{\theta}_k | \fat{u}, \fat{n} )
= 
\mathrm{Dir}( \bldgr{\theta}_k | N_{k,1}+1, \ldots, N_{k,L}+1)
\end{equation*}
with joint histogram counts
\begin{equation}
N_{k,l} = N^0_{k,l} + \sum_i [u_i=l][v_{n_i}=k]
\label{eq:counts_discrete}
.
\end{equation}
Furthermore,
letting
$\bldgr{\Phi}$ be a $I \! \times \! M$ matrix with $\bldgr{\Phi}_i\transp$ in its rows,
letting 
$\bldgr{\delta} = (\delta_1, \ldots, \delta_I)\transp$
collect local, unregularized displacement 
targets
$\delta_i = (n_i - x_i)$,
and 
defining
$\sigma^2 = 9/(8 \pi)$,
we have that
\begin{equation}
p( \fat{c} | \fat{n} ) \simeq \mathcal{N}( \fat{c} | \fat{m}, \fat{S} )
\label{eq:c_posterior}
\end{equation}
with variance
\begin{equation}
\fat{S} 
= 
( \sigma^{-2} \bldgr{\Phi}\transp \bldgr{\Phi} 
+ 
\gamma
\fat{P} )^{-1}
\label{eq:variance}
\end{equation}
and mean
\begin{equation}
\fat{m}
= \sigma^{-2} \fat{S} \bldgr{\Phi}\transp 
\bldgr{\delta}
.
\label{eq:mean}
\end{equation}
This is because a cubic B-spline is 
very
closely approximated by a Gaussian:
$$
\beta( z ) \simeq \mathcal{N}( z | 0, \sigma^2 )
$$
as illustrated in Fig.~\ref{fig:Gaussian_approx}.
As a result,
\begin{eqnarray*}
-\log p( \fat{n} | \fat{c} )
& = &
-\sum_i \log \beta( n_i - y_i ) \\
& \simeq &
\frac{1}{2} \sum_i \frac{( n_i - x_i - \bldgr{\phi}_i\transp \fat{c} )^2}{\sigma^2}
+ \text{const.} \\
& = &
\frac{1}{2\sigma^2}
\| 
\bldgr{\delta}
- \bldgr{\Phi} \fat{c}\|^2
+ \text{const.} 
\end{eqnarray*}
is 
essentially
quadratic in $\fat{c}$.
Since the log-prior on $\fat{c}$ is also quadratic (see \eqref{eq:c_prior}),
the approximation~\eqref{eq:c_posterior} is obtained.

These results 
form the basis of 
a convenient optimization and sampling scheme, 
as detailed below. 


\subsection{Optimization}

Since the node assignments are latent variables in the model, 
it is natural to use an expectation-maximization (EM) algorithm to maximize 
$p(\fat{c}, \bldgr{\theta} | \fat{u})$
with respect to the model parameters.
Starting from initial values
$\fat{c}^{(0)}=\fat{0}$ and 
$\theta_{k,l}^{(0)} = 1/L$,
the 
expected
values of $\fat{n}$ 
at iteration $\tau$
are used
to obtain a new objective function
\begin{equation*}
\Psi^{(\tau)}( \fat{c}, \bldgr{\theta})
=
\sum_{\fat{n}} p(\fat{n} | \fat{u}, \bldgr{\theta}^{(\tau)}, \fat{c}^{(\tau)})
\log p( \fat{c}, \bldgr{\theta} | \fat{u}, \fat{n} )
\end{equation*}
that is then
analytically
optimized to obtain 
updated 
model parameter estimates 
$$
\fat{c}^{(\tau+1)}, \bldgr{\theta}^{(\tau+1)} 
= 
\argmax_{\fat{c}, \bldgr{\theta}}
\Psi^{(\tau)}( \fat{c}, \bldgr{\theta})
$$
at iteration $\tau \! + \! 1$.
It can be shown~\cite{Dempster1977} that each iteration 
of this procedure
increases $\log p(\fat{c}, \bldgr{\theta} | \fat{u})$ by at least the same amount as
$
\Psi^{(\tau)}( \fat{c}, \bldgr{\theta})
$
is increased,
yielding 
a convenient optimization algorithm that does not require tuning.

%
Using the results of the previous section
and
defining
$$
w_{i,j}^{(\tau)} 
= 
p( n_i=j | u_i, \bldgr{\theta}^{(\tau)}, \fat{c}^{(\tau)} )
,
$$
the 
analytical updates for the 
model parameters are given by
$$
\theta_{k,l}^{(\tau+1)}
=
\frac{N_{k,l}^{(\tau)}}{\sum_{l'} N_{k,l'}^{(\tau)}}
,
$$
where 
\begin{equation}
N_{k,l}^{(\tau)} = N_0 + \sum_i \sum_j [u_i=l] [v_j=k] w_{i,j}^{(\tau)}
\label{eq:fractional_counts}
\end{equation}
are expected histogram counts,
and
\begin{equation}
\fat{c}^{(\tau+1)}
= \sigma^{-2} \fat{S} \bldgr{\Phi}\transp 
\bldgr{\delta}^{(\tau)}
\label{eq:Mstep_c}
\end{equation}
where
$\delta_i^{(\tau)} = \sum_j w_{i,j}^{(\tau)} j - x_i$
are
expected local displacement 
targets.

Note that the fractional histogram counts of~\eqref{eq:fractional_counts} 
are very similar to those arising in \emph{partial volume interpolation}
~\cite{Maes1997,Chen2003}, where
effectively
the prior
$
p( n_i = j | \fat{c}^{(\tau)} )
$
-- instead of the posterior --
is used to distribute each voxel's contribution.



\subsection{Sampling}

%
A very similar approach can be used to construct a tuning-free MCMC sampler.
Again 
starting from some initial values $\fat{c}^{(0)}$ and $\bldgr{\theta}^{(0)}$,
we can sample from $p(\fat{c}, \bldgr{\theta}, \fat{n} | \fat{u})$ 
by
drawing 
from the conditional distribution of each of the three variables in turn (Gibbs sampler):
\begin{eqnarray*}
n_i^{(\tau + 1)} & \sim &
\mathrm{Cat}( \ldots, w_{i,j}^{(\tau)}, \ldots ),
\quad \forall i
\\
\bldgr{\theta}_k^{(\tau + 1 )} & \sim &
\mathrm{Dir}( \ldots, N_{k,l}^{(\tau+1)}\!\!+1, \ldots ),
\quad \forall k
\\
\fat{c}^{(\tau + 1 )} & \sim & 
\mathcal{N}( \fat{m}^{(\tau + 1 )}, \fat{S} )
.
\end{eqnarray*}
Here
$N_{k,l}^{(\tau+1)}$ 
and $\fat{m}^{(\tau + 1 )}$ 
are given by~\eqref{eq:counts_discrete} and~\eqref{eq:mean} when the node assignments are $\fat{n}^{(\tau + 1)}$.

After running the resulting chain for $T$ iterations and discarding the first $T_0$ ones
-- when the chain is still converging to the target distribution --
the set 
$\{ \fat{c}^{(\tau)} \}_{\tau=T_0+1}^{T}$ will contain $(T-T_0)$ correlated samples from the marginal distribution $p(\fat{c}| \fat{u})$.

%
%
%
%
%

\subsection{Application to nonlinear registration}

Although the proposed methods are generic and can be used with other basis functions and regularizers,
here we concentrate on nonlinear registration 
with discrete cosine transform (DCT) basis functions
and bending energy regularization.
This results in an efficient, demons-like algorithm in which local target displacements $\bldgr{\delta}$ are iteratively estimated and 
spatially 
filtered
(with reflexive boundary conditions)
to obtain 
regularized deformation fields.

In 
detail, when the columns of $\bldgr{\Phi}$ are the $M\!=\!I$ basis functions of the orthonormal DCT, 
we have that
$\bldgr{\Phi}\transp \bldgr{\Phi} \!=\! \fat{I}_I$.
Furthermore, as detailed in Sec.~VIII of the supplementary material,
penalizing second-order derivatives results 
in a diagonal precision matrix $\fat{P}$,
%
so that
$\fat{S}$ is also diagonal (see~\eqref{eq:variance}).
Therefore,
each EM iteration involves
one DCT (to compute 
$\bldgr{\Phi}\transp \bldgr{\delta}^{(\tau)}$ in~\eqref{eq:Mstep_c}),
followed by an
element-wise multiplication 
attenuating
high frequencies 
(multiplying by $\sigma^{-2} \fat{S}$
to obtain $\fat{c}^{(\tau+1)}$ in~\eqref{eq:Mstep_c}), 
followed by an inverse DCT 
(to obtain the resulting deformation field $\fat{\Phi} \fat{c}^{(\tau+1)}$).
The sampler 
follows a similar pattern, but
adds element-wise Gaussian noise to each coefficient before
the inverse DCT.
Both the DCT and inverse DCT can be computed in $\mathcal{O}(I \log I)$ time using FFTs.

\if\hideResults0

\section{Optimization Experiments and Results}



To compare the performance of the optimization variant of the proposed method against that of several state-of-the-art techniques for nonlinear registration,
we 
evaluated 
both the registration accuracy of each method
and 
how sensitive it is
to the tuning of its hyperparameters.
Towards this end, we 
used each method to register
over
10,000 image pairs
taken from
three 
different anatomical regions (abdomen, lung and brain).
These image pairs
cover
both inter- and intra-subject registration 
across various
inter- and intra-modality 
scenarios,
as illustrated in Fig.~\ref{fig:datasets}.
%

\subsection{Registration tasks} 


We combined publicly available training data of the Learn2Reg2022 challenge~\cite{Hering2022}\footnote{\url{https://learn2reg.grand-challenge.org/}}
with data from the Open Access Series of Imaging Studies (OASIS-3) project~\cite{Lamontagne2019} \footnote{\url{https://www.oasis-brains.org/}} to define 
the following 9 registration tasks\footnote{We use the convention that the first and second image of an image pair correspond to the fixed and moving image, respectively.}:






\begin{itemize}

    \item[-] A first task 
    consists of registering abdominal MR scans of 30 subjects 
    with abdominal CT scans of 30 other subjects.
    The data originates from Learn2Reg2022 and underwent several pre-processing steps performed by the challenge organizers, including
    affine registration, resampling to 2mm isotropic resolution, and masking of a relevant region within the images~\cite{Clark2013,Xu2016,Kavur2021}.
    We evaluated registration accuracy by computing the Dice score between manual 
    annotations of the liver, the spleen, the left kidney and the right kidney
    after nonlinear registration. 

    %
    %
    
    \item[-] A second 
    task
    is 
    very 
    similar to the first one, but uses abdominal CT scans of 20 subjects for intra-modality registration instead. This task's data also originates from Learn2Reg2022, with the same type of preprocessing~\cite{Xu2016}. 

    
    \item[-] A third task performs intra-subject registration between lung CT scans acquired at two different screenings in 80 subjects with lung cancer. 
    This data too originates from Learn2Reg2022, 
    and comes with masks delineating the lungs as well as automatically annotated landmarks~\cite{Heinrich2015} that establish the correspondence between each subject's initial screening and their follow-up. 
    We evaluated registration accuracy by computing the
    target registration error (TRE), defined as the average Euclidean distance between corresponding landmarks 
    in each subject
    after nonlinear registration.
    
    %
    
    \item[-] 
    Our remaining six tasks consist of inter-subject registration of T1-weighted, T2-weighted, and FLAIR brain MR scans 
    between
    39 subjects, 
    in particular the combinations 
    T1-T1, T1-T2, T2-T2, T1-FLAIR, T2-FLAIR and FLAIR-FLAIR.
    The images are part of the OASIS-3 project, and were acquired with three different Siemens scanners with multiple field strengths 
    at various resolutions.
    Registration accuracy was evaluated by computing the average Dice overlap across 13 brain structures that were automatically segmented from each subject's multicontrast scan (i.e., T1\,\,+\,\,T2\,\,+\,\,FLAIR) using the SAMSEG~\cite{Puonti2016} tool distributed with FreeSurfer. In particular, 
    Dice scores were computed for the following structures: 
    brain stem, cerebellum white matter, cerebral white matter, cerebellum cortex, cerebral cortex, amygdala, hippocampus, nucleus accumbens, caudate, lateral ventricle, putamen, pallidum and thalamus, with left-right scores averaged.
    %
    For each registration experiment, the input scans consisted of bias field corrected images produced by SAMSEG, which are defined within a mask surrounding the head. The nonlinear registration of each image pair was initialized using an affine registration computed with FLIRT~\cite{Jenkinson2002}.

\end{itemize}

Since each of the 
methods we compare against 
can be tuned to 
optimize
performance 
in a particular setting,
we 
also
composed
a matched validation dataset
for each registration task.
This validation set consisted of the images of 10 (for the abdomen CT-CT and the six brain tasks) or 20 (for abdomen MR-CT and lung CT-CT) additional subjects that were distinct from the ones used to evaluate accuracy.

%
%
%
%
%
%
%
%

\begin{figure*}[!ht]
    \centering
    \includegraphics[width=0.88\linewidth]{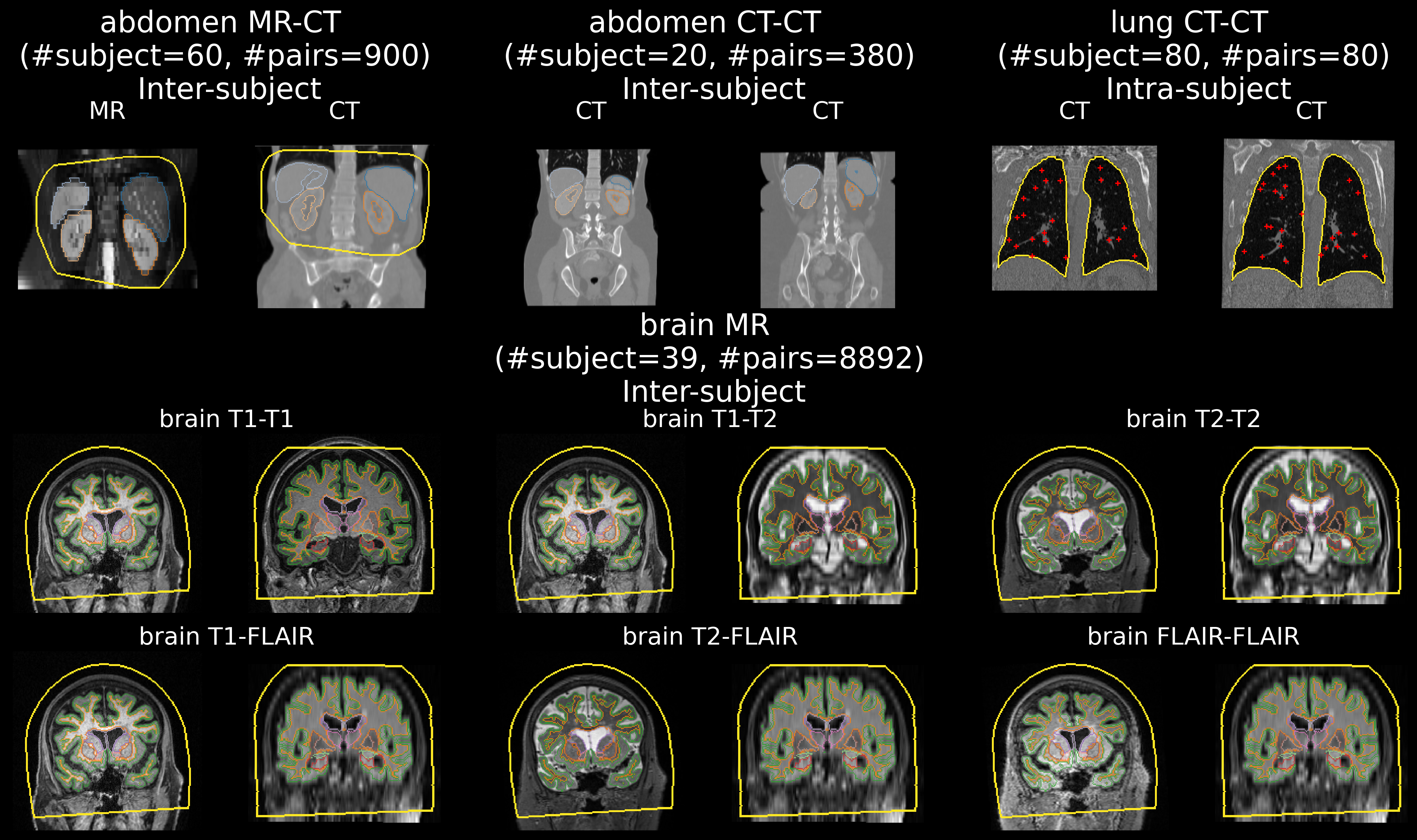}
    \caption{
    Representative images for each of the abdomen MR-CT, abdomen CT-CT, lung CT-CT, and brain MR datasets. Segmentation overlays are shown in color where available, and yellow contours indicate image masks. For NLST, available keypoints are displayed with a red plus sign. 
    }
    \label{fig:datasets}
\end{figure*}

\subsection{Benchmark methods}


Although 
our 9 registration tasks 
can 
easily
be 
solved
by training 
dedicated learning-based solutions,
here 
we 
are interested in 
methods that 
work 
robustly
across a wide range of tasks, 
without access to task-specific training data.
%
We therefore benchmarked against three well-established, publicly available toolboxes and one domain-randomized learning-based method:
\begin{list}{-}{}

    \item \textbf{Elastix}\footnote{\url{https://github.com/SuperElastix/elastix}}, 
    a general-purpose image registration toolbox, widely used for its B-spline free-form deformations and its adaptive stochastic gradient descent optimizer~\cite{Klein2009,Shamonin2014}.
    
    \item \textbf{NiftyReg}\footnote{\url{https://github.com/KCL-BMEIS/niftyreg}},
    a flexible command-line registration tool that also optimizes cubic B-spline free-form deformations, often chosen for its computational speed and robust default parameters~\cite{Rueckert1999,Modat2010}. 
    
    \item \textbf{ANTs}\footnote{\url{https://github.com/ANTsX/ANTs}},
    a registration toolbox that is particularly used in neuroimaging for cross-subject brain registration, because of its highly flexible symmetric diffeomorphic deformation model~\cite{Avants2008, Tustison2013}.
    
    \item \textbf{SynthMorph}\footnote{\url{https://github.com/voxelmorph/voxelmorph}}, 
    a learning-based method trained purely on image pairs synthesized from label maps of either randomly generated shapes or anatomical segmentations~\cite{Hoffmann2022}.
        
\end{list}
Even though all these tools 
are task-agnostic,
each of 
them
also
allows the user to tweak some of its 
settings 
to optimize 
performance. 
The range of settings that we considered in our experiments, along with 
more detailed information about each method's software implementation in general, are provided in
Sec.~IX of the supplementary material.


\subsection{Implementation of our algorithm}


We implemented the proposed optimization algorithm in an open-source Python library\footnote{\url{https://github.com/ste93ste/BINDER}} that wraps C++ and CUDA code on CPU and GPU, respectively.
In our implementation, the input image intensities are cropped within the 1\% and 99\% percentiles and then discretized into $K=L=64$ equally spaced bins.
If a mask for an input image is provided, the intensities outside of the mask are simply set to zero;
the moving image is always virtually padded with zeros to have infinite support.

%
We employ a multi-resolution optimization scheme with four levels, where the number of voxels is reduced by a factor of 
2 in each spatial dimension
between the levels. 
When the algorithm has converged at one lower-resolution level (detected as a voxel-wise change in objective function of 
$5 \cdot 10^{-5}$ 
or less),
the computed deformation field is upsampled using linear interpolation to initialize the next level. 


Since the deformation fields computed by 
ANTs and SynthMorph are 
designed to be
diffeomorphic (topology preserving and smooth)
%
whereas ours are not, these methods may be disadvantaged when only registration overlap is assessed.  
To address this issue, 
we 
followed
~\cite{Karaccali2004, Andersson2019}
and projected, at the end of each multiresolution level, 
the computed deformation field onto 
the
closest diffeomorphic one
with
Jacobian determinants 
in
the range $[0.01, 100]$.
%
%
%
%
%
%
%
%
%
%
%
The effect of applying this projection procedure is illustrated in Table~\ref{tab:topologyCorrection}
for 
the validation dataset.
%
Across all tasks except one, the deformation fields produced by the original algorithm are already very close to topology preserving, so that 
their projection
does not 
change
the 
registration accuracy 
metrics.
The one exception is the abdomen CT-CT task, where 
the average Dice score is reduced by 0.019 when the deformations are 
forced to become
diffeomorphic.
The results reported in the remainder of the paper were all obtained with the projection procedure switched on.


\begin{table}[!t]
\begin{center}
\resizebox{0.37\textwidth}{!}{%
\iffalse
\begin{tabular}{|c||c|c||c|c|}
\hline
\multirow{2}{*}{Dataset} & \multicolumn{2}{c||}{Dice $\uparrow$ (*TRE $\downarrow$)} & \multicolumn{2}{c|}{\# det(Jacobians) \textless 0 $\downarrow$} \\ 
\cline{2-5} 
& TopCor & NoTopCor & TopCor & NoTopCor \\ 
\hline
abdomen MR-CT 
& \textbf{0.641} & \textbf{0.641}
& \textbf{0.0} & 0.48 \% 
\\ 
\hline
abdomen CT-CT 
& 0.501 & \textbf{0.520} 
& \textbf{0.0} & 2.12 \% 
\\
\hline
lung CT-CT 
& \textbf{0.93*} & \textbf{0.93*} 
& \textbf{0.0} & $<$ 0.01 \% 
\\ \hline
brain T1-T1  
& \textbf{0.797} & \textbf{0.797} 
& \textbf{0.0} & 0.15 \% 
\\ \hline
brain T1-T2 
& \textbf{0.732} & \textbf{0.732} 
& \textbf{0.0} & 0.11 \% 
\\ \hline
brain T2-T2 
& \textbf{0.755} & \textbf{0.755} 
& \textbf{0.0} & 0.10 \% 
\\ \hline
brain T1-FLAIR 
& \textbf{0.697} & \textbf{0.697}
& \textbf{0.0} & 0.02 \% 
\\ \hline
brain T2-FLAIR 
& \textbf{0.680} & \textbf{0.680} 
& \textbf{0.0} & $<$ 0.01 \% 
\\ \hline
brain FLAIR-FLAIR 
& \textbf{0.744} & \textbf{0.744} 
& \textbf{0.0} & 0.03 ;\% 
\\
\hline
\end{tabular}
\else
\begin{tabular}{|l||c|c||c|c|}
\hline
\multirow{2}{*}{} & 
\multicolumn{2}{c||}{\# det(Jacobians) \textless 0 $\downarrow$} &
\multicolumn{2}{c|}{Dice $\uparrow$ (*TRE $\downarrow$)} \\
\cline{2-5} 
& NoProj & Proj & NoProj & Proj \\ 
\hline
brain T1-T1  
& 0.15 \% 
& 0.0 
& 0.797 
& 0.797 
\\ \hline
brain T1-T2 
& 0.11 \% 
& 0.0 
& 0.732 
& 0.732 
\\ \hline
brain T2-T2 
& 0.10 \% 
& 0.0 
& 0.755 
& 0.755 
\\ \hline
brain T1-FLAIR 
& 0.02 \% 
& 0.0 
& 0.697 
& 0.697
\\ \hline
brain T2-FLAIR 
& $<$ 0.01 \% 
& 0.0 
& 0.680 
& 0.680 
\\ \hline
brain FLAIR-FLAIR 
& 0.03 \% 
& 0.0 
& 0.744 
& 0.744 
\\
\hline
abdomen MR-CT 
& 0.48 \% 
& 0.0 
& 0.641
& 0.641 
\\ 
\hline
abdomen CT-CT 
& 2.12 \% 
& 0.0 
& 0.520 
& 0.501 
\\
\hline
lung CT-CT 
& $<$ 0.01 \% 
& 0.0 
& 0.93* 
& 0.93* 
\\ \hline
\end{tabular}
\fi
} 
\end{center}
\caption{
%
Effect of the projection procedure to make the obtained deformation fields diffeomorphic, evaluated on the validation dataset. For each task, the $\gamma$ hyperparameter was set to the value yielding the highest accuracy metric. 
%
The proportion of voxels with negative Jacobian determinants is averaged over registration pairs; 
accuracy metrics are also averaged over structures. 
Proj = Projection; NoProj = No Projection. 
}
\label{tab:topologyCorrection}
\end{table}


%

%
Typical running times of a complete nonlinear registration experiment
range from 2.63 minutes for the lung CT-CT task to 8.49 minutes for the abdomen CT-CT task
on CPU 
(Intel Core i9-13900K processor with 32 threads).
%
On GPU 
(NVIDIA H100 with 80 GB of memory)
this 
reduces
to around 22 seconds across all tasks.

\subsection{Results}



We report two sets of results regarding the 
accuracy 
obtained by
the various methods on the 9 registration tasks.
The first one corresponds to the way registration algorithms are typically evaluated in the literature: for each task, the user settings of each algorithm were first tuned on the corresponding task-specific held-out images of the validation set. This was performed using a grid search over the range of hyperparameter values considered for each method, as detailed in Sec.~IX of the supplementary material. Once the tuning achieving the highest registration accuracy in the validation set was determined for a particular task, it 
was then used to compute the performance on 
that task in the 
test data.
%

%
In addition to 
these
results, which 
only
assess \emph{peak accuracy},
we also computed a second set of results that aim to evaluate the \emph{sensitivity} of each method to the tuning of its hyperparameters. 
In particular, 
we re-ran each method on the 9 tasks, but 
for each registration pair we now randomly selected the tuned configuration setting for one of the 8 \emph{other} tasks, rather than for the task itself.
With this set-up we aimed to simulate a real-world clinical scenario in which registration needs to be performed in a new application area: 
In such cases dedicated annotated validation data for algorithmic tuning does not typically exist, 
but a library of working configurations from other, existing applications will often be available.
The experiment therefore tries to quantify how well the various methods 
may
work directly ``out-of-the-box''
in the hands of end-users,
without prior expert tuning.


The results of both these ``tuned'' and ``untuned'' experiments are summarized in Tables~\ref{tab:tuned} and~\ref{tab:untuned}, respectively, with a visual representation
in  
Fig.~\ref{fig:tuned_vs_untuned}. 
In the figure, 
average 
tuned
accuracy can be read off on the horizontal axis in the scatter plots, 
whereas sensitivity to hyperparameter tuning is reflected in deviations from the identity line;
an ideal method would 
have results 
that fall exactly on the line and are
concentrated in the upper right corner.
In general, 
the proposed method 
seems to work 
the most
robustly
across all structures and in both the tuned and untuned scenarios. In non-brain structures it typically outperforms the tuned benchmarks even when untuned; in the brain it is 
often
the better method when untuned, and a close second (together with ANTs) when tuned. 
NiftyReg, on the other hand, is the best tuned method for the brain, but it generally underperforms in non-brain structures, and it appears to be very sensitive to the tuning of its hyperparameters.
In comparison,
the 
tuned 
peak performance of ANTs 
often
does not 
fully reach 
that of NiftyReg, 
but it is also less sensitive to its tuning;
SynthMorph and Elastix generally are not the top performers in either the tuned or the untuned setting.

\begin{table}
\scriptsize
\begin{tabular}{|l|c|c|c|c|c|}
\hline
 & SynthMorph & NiftyReg & ANTs & Elastix & Ours\\
\hline
brain T1-T1  $\uparrow$ & 0.777 & \bf{0.798} & 0.794 & 0.771 & 0.791\\
\hline
brain T1-T2  $\uparrow$ & 0.571 & 0.699 & 0.604 & 0.552 & \bf{0.726}\\
\hline
brain T2-T2  $\uparrow$ & 0.705 & 0.759 & \bf{0.760} & 0.710 & 0.759\\
\hline
brain T1-FLAIR  $\uparrow$ & 0.665 & \bf{0.716} & 0.708 & 0.633 & 0.708\\
\hline
brain T2-FLAIR  $\uparrow$ & 0.699 & \bf{0.706} & 0.673 & 0.604 & 0.689\\
\hline
brain FLAIR-FLAIR  $\uparrow$ & 0.728 & \bf{0.745} & 0.744 & 0.689 & 0.737\\
\hline
abdomen MR-CT  $\uparrow$ & 0.254 & 0.314 & 0.327 & 0.407 & \bf{0.754}\\
\hline
abdomen CT-CT  $\uparrow$ & 0.552 & 0.669 & 0.628 & 0.665 & \bf{0.755}\\
\hline
lung CT-CT  $\downarrow$ & 2.066 & \bf{0.991} & 1.436 & 3.874 & 1.044\\
\hline
\end{tabular}
\caption{Registration accuracy of the various methods on the 9 tasks when their settings have been tuned on well-matched, task-specific validation image pairs. 
Accuracy metrics (TRE for lung CT-CT, Dice for all other tasks) have been averaged across structures and registration pairs.
Boldface indicates the ``winning'' method for each task.
}
\label{tab:tuned}
\end{table}

\begin{table}[!t]
\scriptsize
\begin{tabular}{|l|c|c|c|c|c|}
\hline
 & SynthMorph & NiftyReg & ANTs & Elastix & Ours\\
\hline
brain T1-T1  $\uparrow$ & 0.767 & 0.770 & \bf{0.780} & 0.738 & 0.775\\
\hline
brain T1-T2  $\uparrow$ & 0.555 & 0.624 & 0.447 & 0.427 & \bf{0.713}\\
\hline
brain T2-T2  $\uparrow$ & 0.714 & 0.740 & 0.751 & 0.698 & \bf{0.752}\\
\hline
brain T1-FLAIR  $\uparrow$ & 0.666 & 0.667 & 0.616 & 0.588 & \bf{0.686}\\
\hline
brain T2-FLAIR  $\uparrow$ & \bf{0.687} & 0.649 & 0.496 & 0.525 & 0.650\\
\hline
brain FLAIR-FLAIR  $\uparrow$ & 0.715 & \bf{0.731} & 0.722 & 0.678 & 0.731\\
\hline
abdomen MR-CT  $\uparrow$ & 0.182 & 0.264 & 0.301 & 0.336 & \bf{0.739}\\
\hline
abdomen CT-CT  $\uparrow$ & 0.536 & 0.508 & 0.577 & 0.602 & \bf{0.677}\\
\hline
lung CT-CT  $\downarrow$ & 2.788 & 2.561 & 1.821 & 4.297 & \bf{1.075}\\
\hline
\end{tabular}
\caption{
Similar to
Table~\ref{tab:tuned}, but this time when a 
configuration setting optimized for one of the other tasks was randomly selected.
}
\label{tab:untuned}
\end{table}

\begin{figure*}[!ht]
  \setlength{\mywidth}{0.36\linewidth}
  \centering
  \begin{tabular}{cc}
    \includegraphics[width=\mywidth]{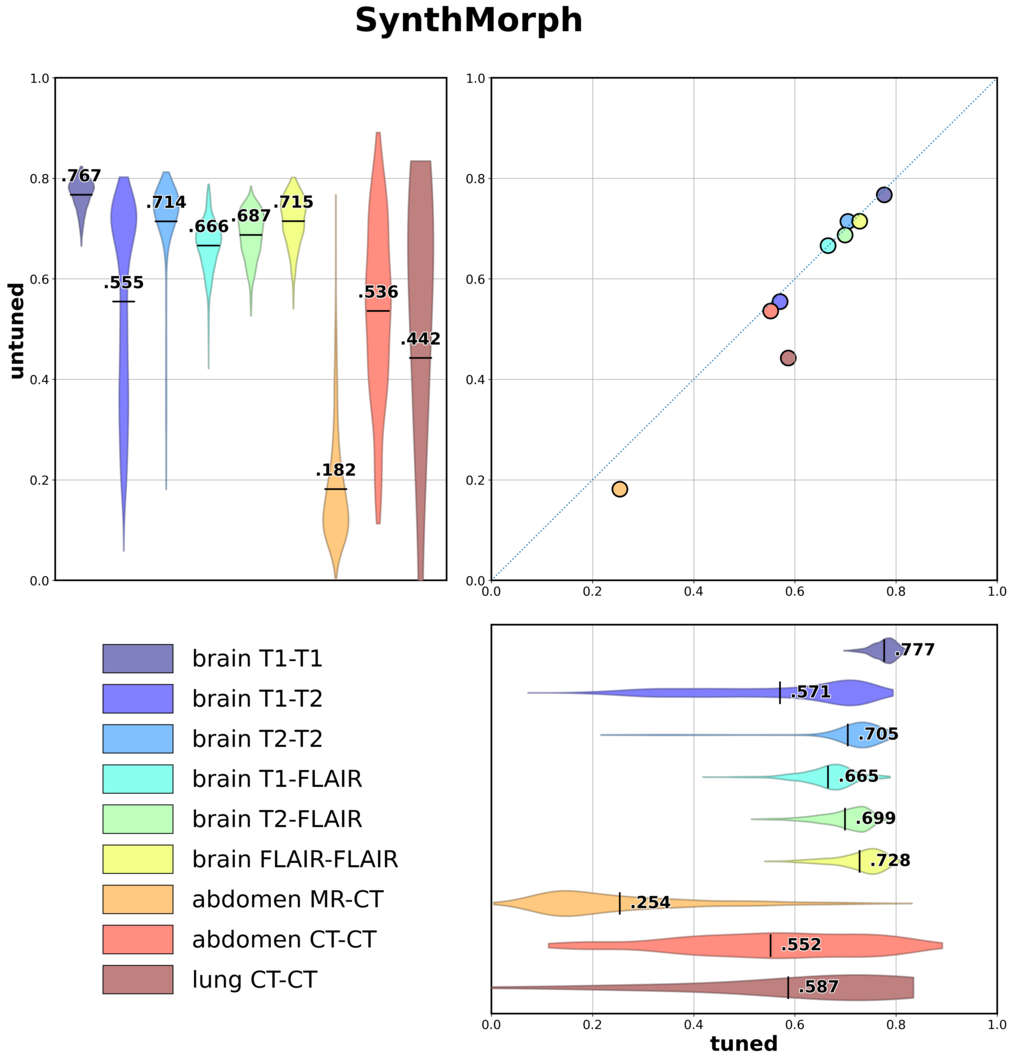} &
    \includegraphics[width=\mywidth]{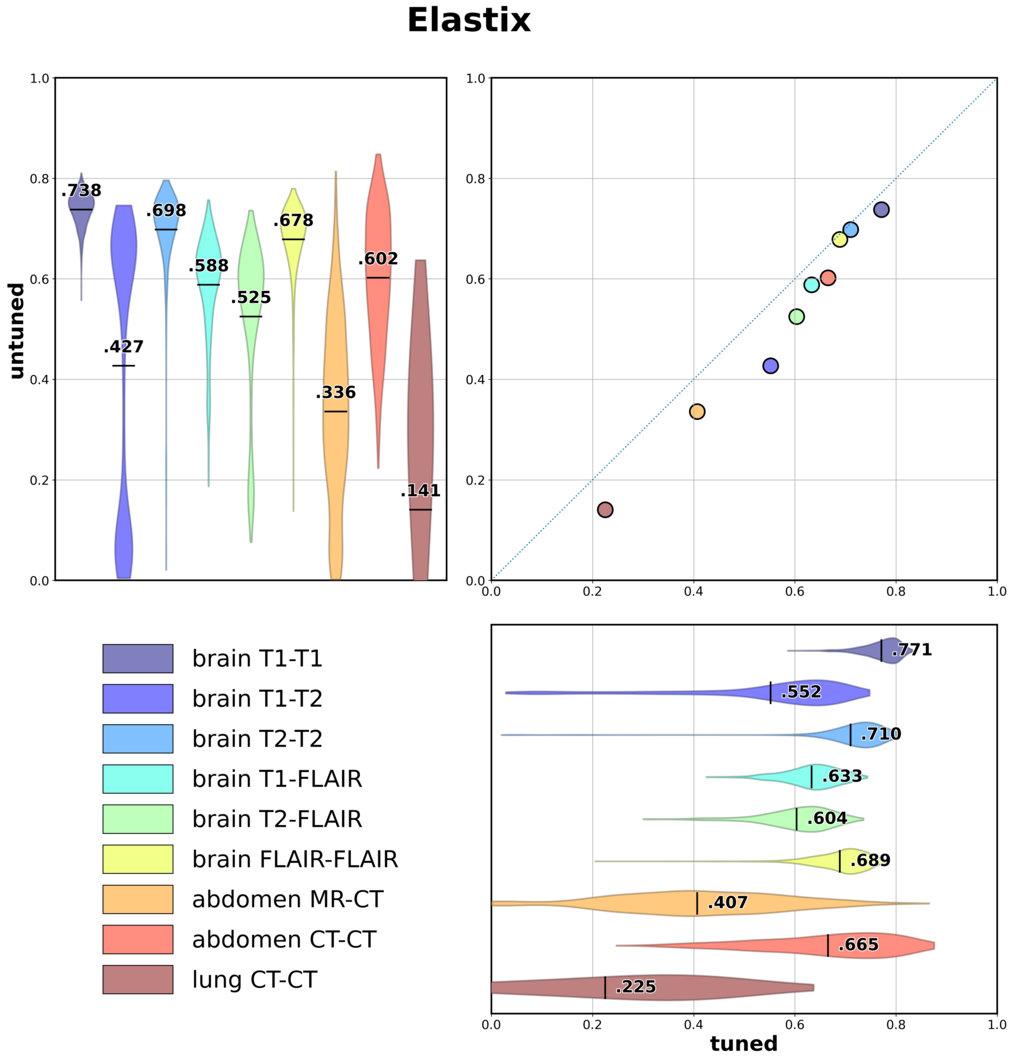} \\
    \includegraphics[width=\mywidth]{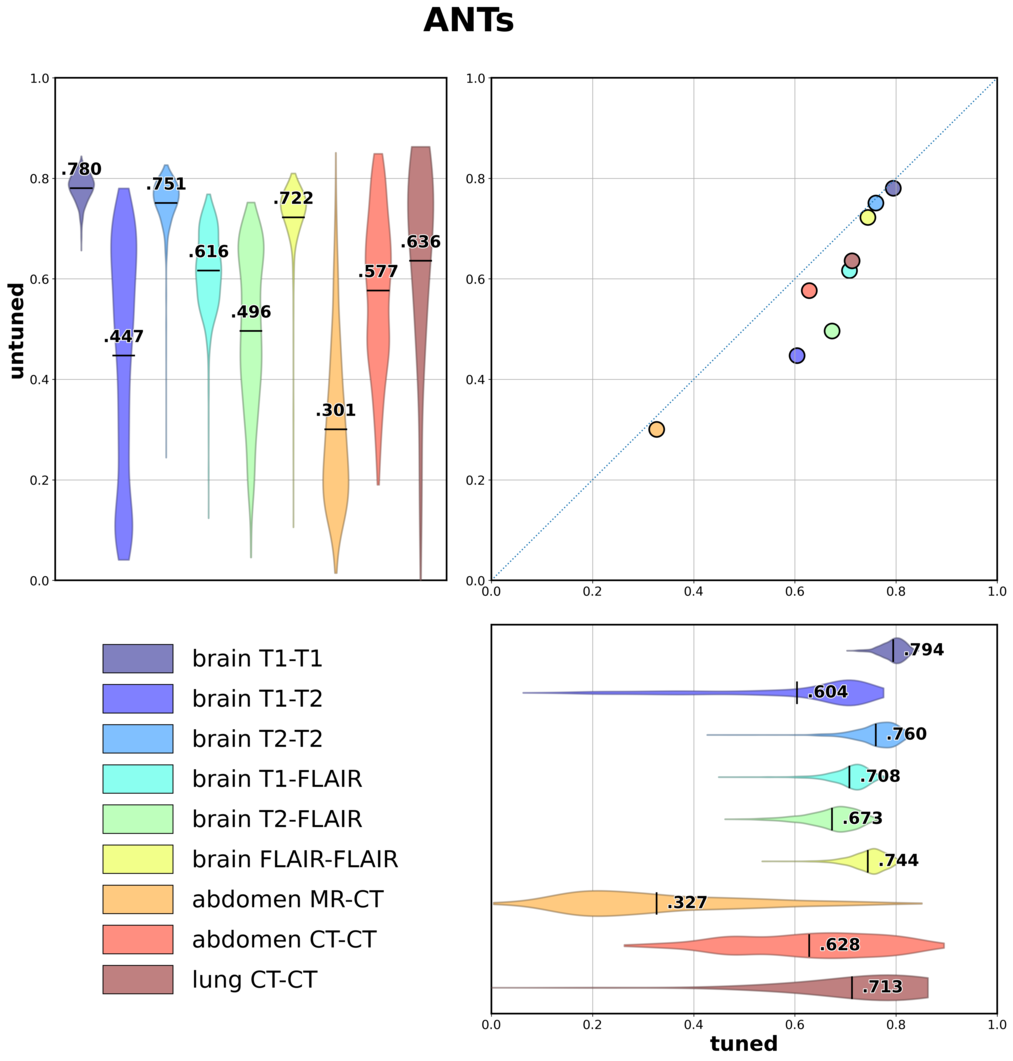} &
    \includegraphics[width=\mywidth]{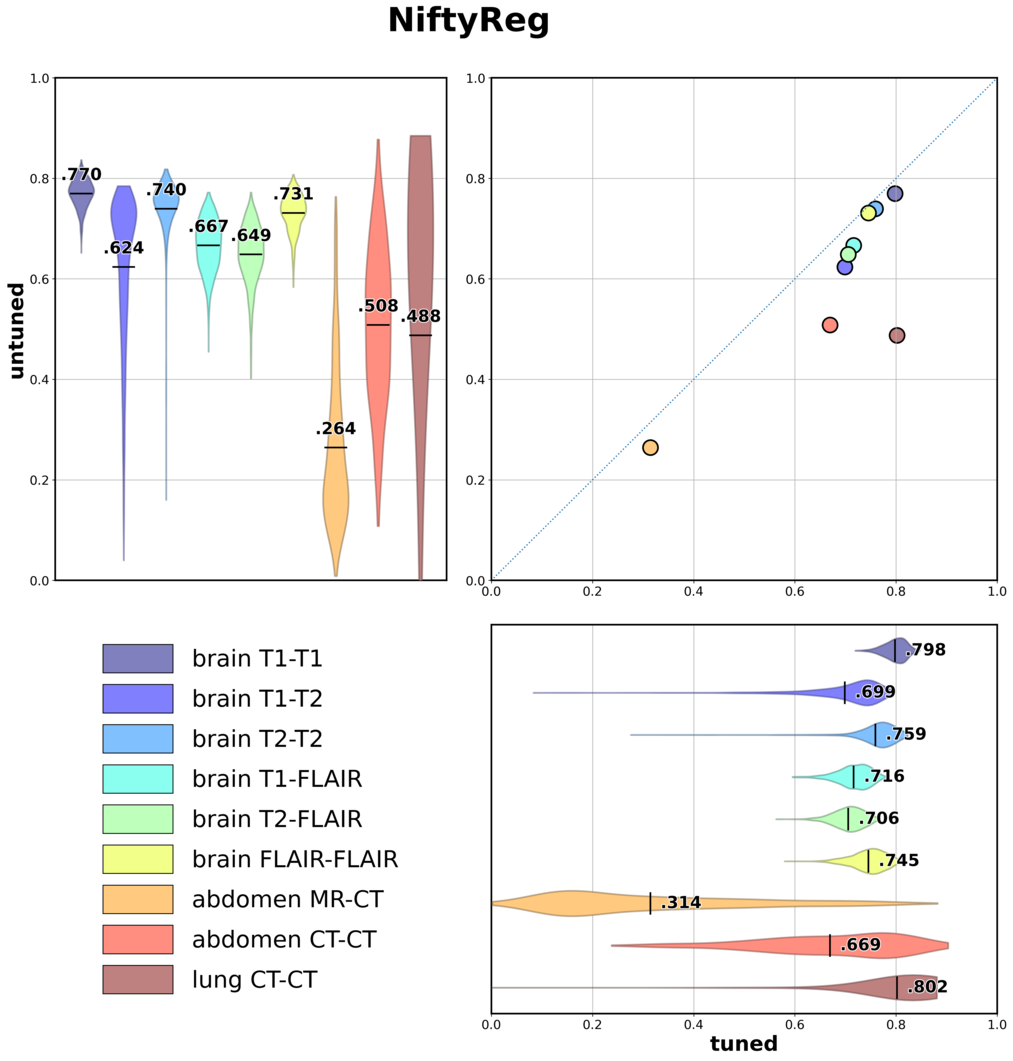} \\
    \includegraphics[width=\mywidth]{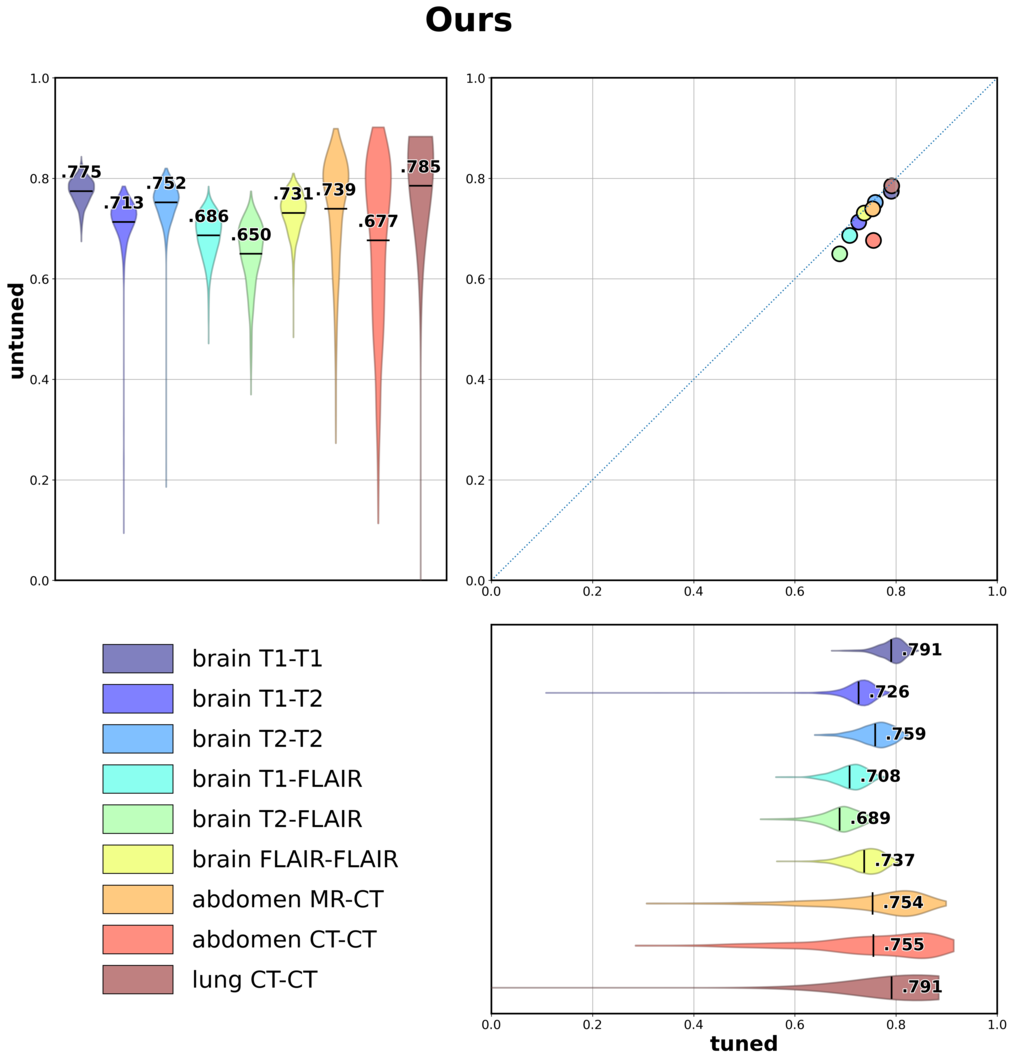} & 
    \begin{minipage}[b]{\mywidth}
      \caption{
      Visual representation of the results summarized in Tables~\ref{tab:tuned} and~\ref{tab:untuned}.
      Each scatter plot shows the average Dice performance of a method when it is tuned (horizontal axis) vs.~when it is not 
      (vertical axis). 
      Ideally both should attain the same high values. 
      For each task, violin plots summarize the results across all registration pairs, with the black line indicating the mean. 
      Since the lung CT-CT task is evaluated using TRE instead of Dice, a pseudo-Dice score was computed as
      $-TRE/5+1$, just so
      that its results could also be visualized in the same plots.
      }
      \label{fig:tuned_vs_untuned}
    \end{minipage}
    \\
  \end{tabular}
\end{figure*}


\section{Sampling Experiments and Results
}
\label{sec:sampling}

We illustrate the proposed Gibbs sampler on inter-subject brain registration using images from the OASIS-3 dataset, both in 
multimodal (T1-T2) and monomodal (T1-T1) settings.
A summary of the results for two such cases is shown in Figs~\ref{fig:case4} and~\ref{fig:case2}, 
with additional results on two more cases in Sec.~X of the supplementary material. The supplementary material also contains, for each of the four cases, additional visualizations
%
%
with individual deformation samples and spatial maps illustrating the convergence properties of the sampler.

\setlength{\myheight}{0.25\textwidth}
\setlength{\myspace}{1ex}

\begin{figure*}[!ht]
  \centering
  \begin{tabular}{c@{\hspace{\myspace}}c@{\hspace{\myspace}}c}
    \includegraphics[height=\myheight]{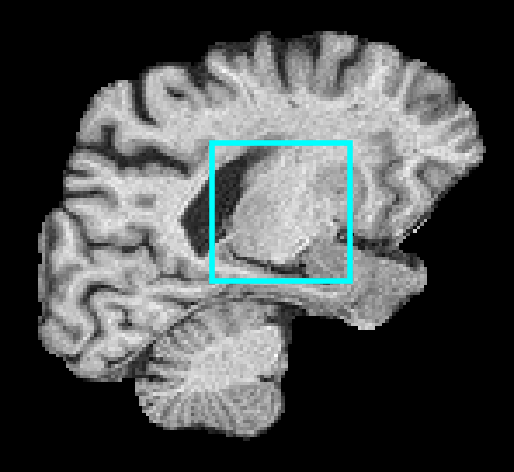} & 
    \includegraphics[height=\myheight]{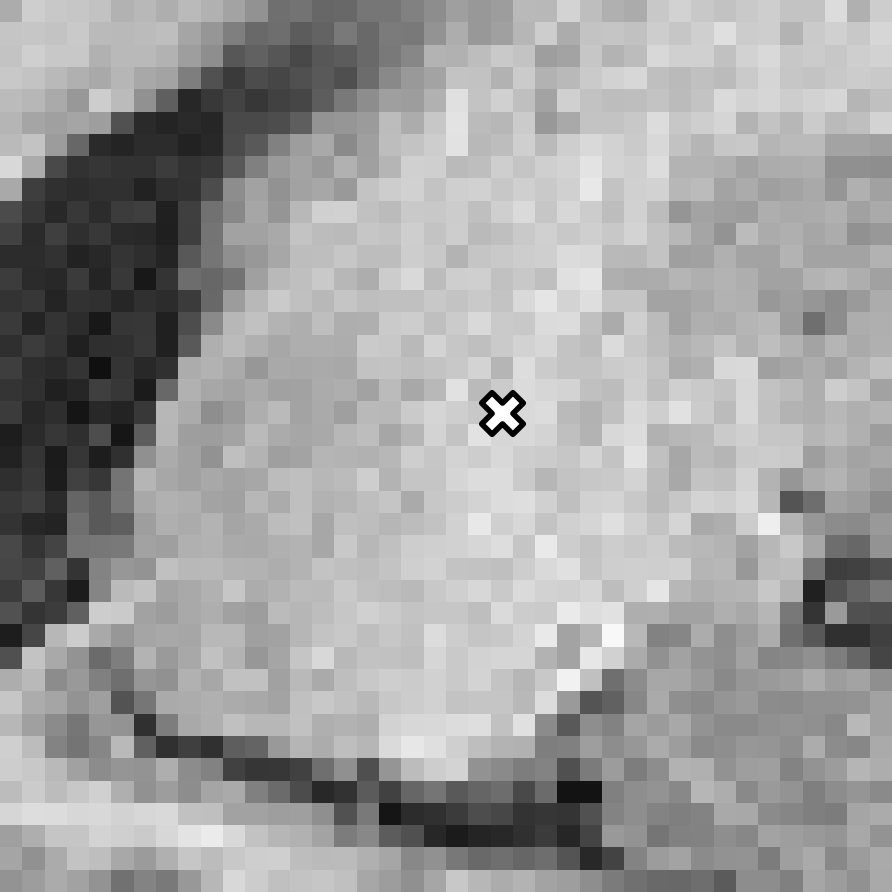} &
    \includegraphics[height=\myheight]{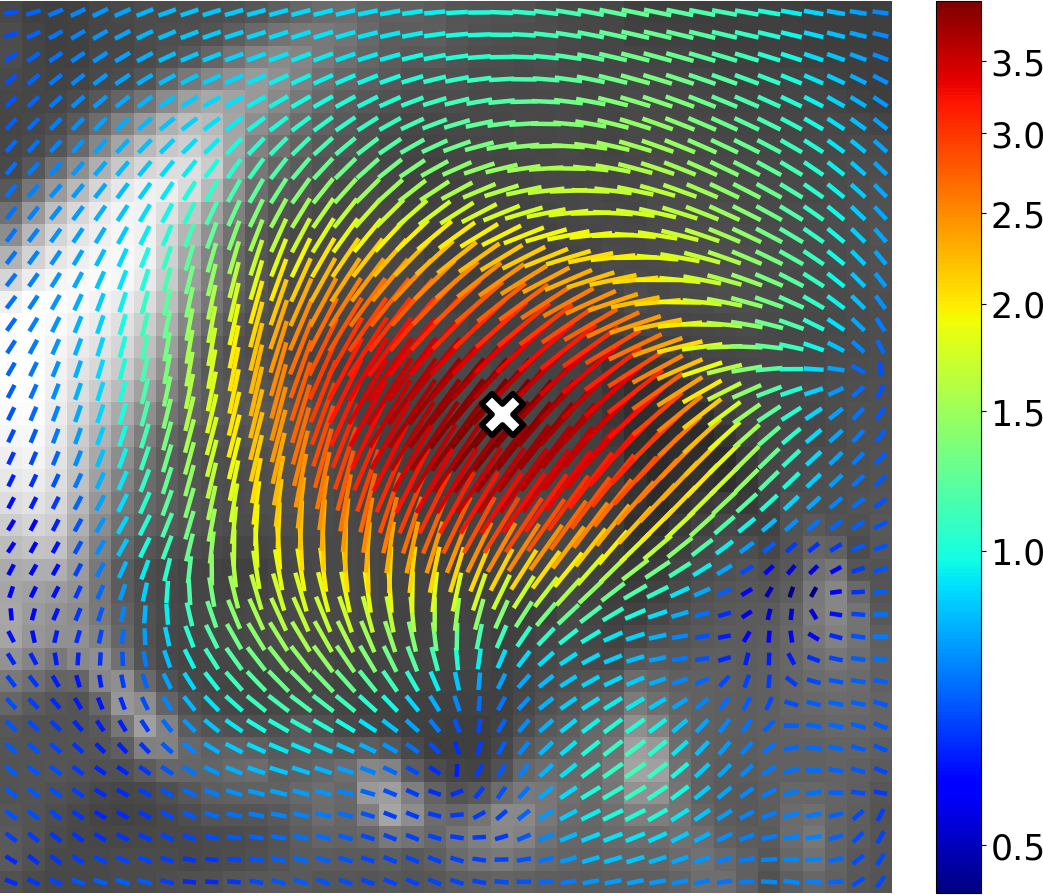} \\   
    \includegraphics[height=\myheight]{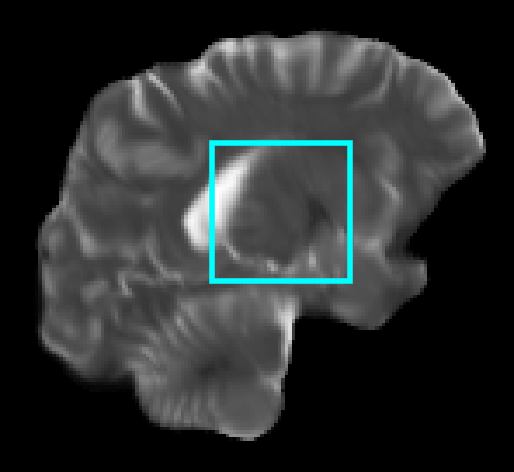} &
    \includegraphics[height=\myheight]{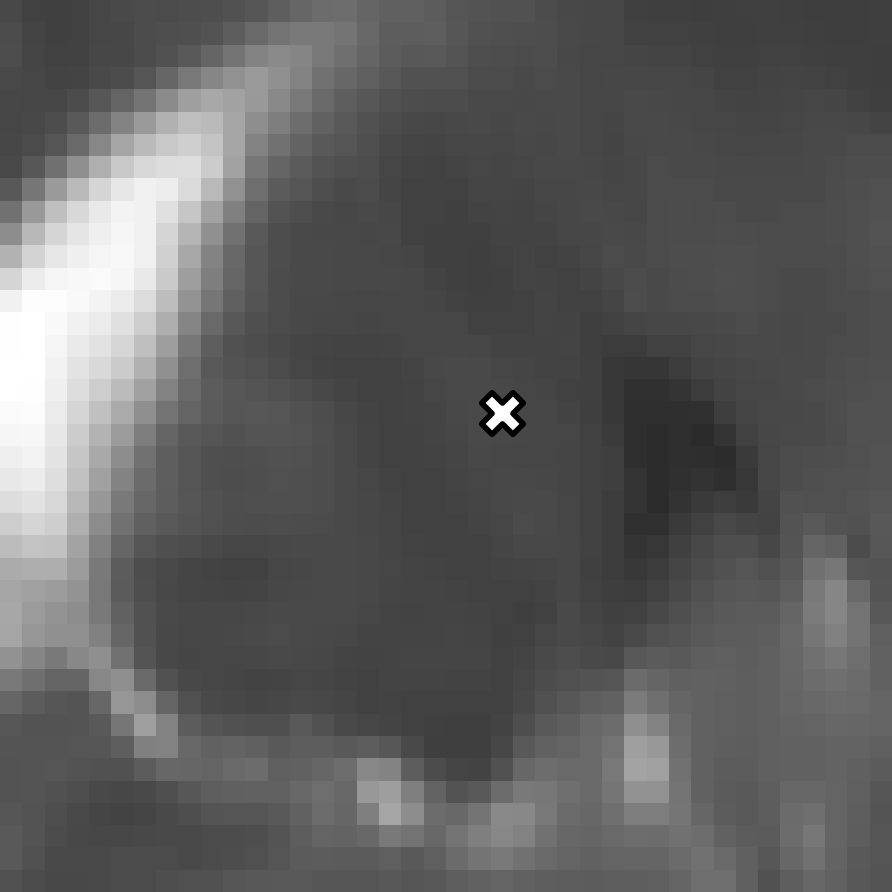} &          
    \includegraphics[height=\myheight]{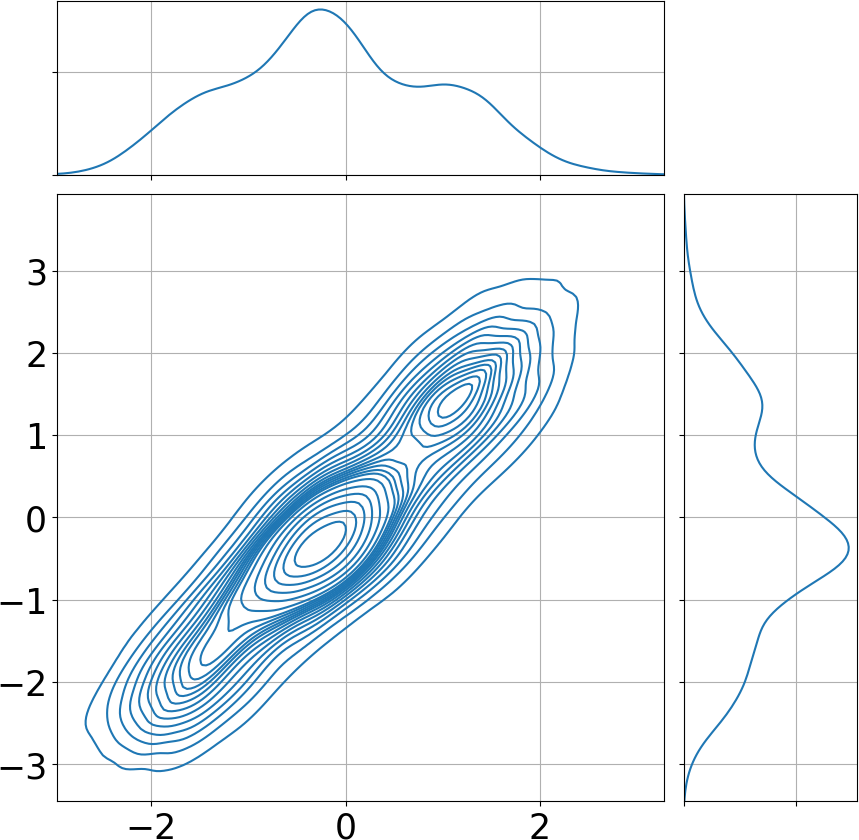}
  \end{tabular}
  \caption{Uncertainty quantification with the proposed Gibbs sampler in an inter-subject brain registration example involving different modalities (T1-weighted vs.~T2-weighted MRI scans). The left column shows the fixed image (top) and the resampled moving image (bottom) after deforming it with the mean deformation across 11 million samples. The middle column shows the same but zoomed in to the ROI indicated by the white square in the left column. The top figure in the right column visualizes the largest eigenvector of the $2 \times 2$ covariance matrix in each voxel (the dominant direction of disagreement across the samples),
  centered around that voxel and 
  scaled by two times the square root of the corresponding eigenvalue (to indicate both directions).
  %
  Finally, the bottom figure in the right column shows the 2D marginal distribution around the mean deformation in the voxel indicated by the white cross in the zoomed-in plots. This distribution was estimated from the 11 million samples using kernel density estimation.
  The units are measured in voxels.
  }
  \label{fig:case4}
\end{figure*}  

\begin{figure*}[!ht]
  \centering
  \begin{tabular}{c@{\hspace{\myspace}}c@{\hspace{\myspace}}c}
    \includegraphics[height=\myheight]{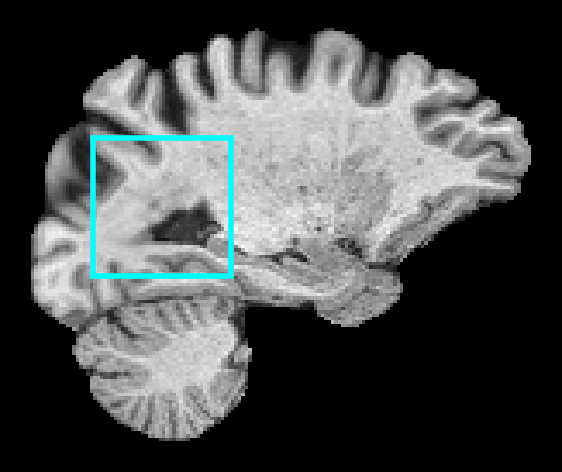} & 
    \includegraphics[height=\myheight]{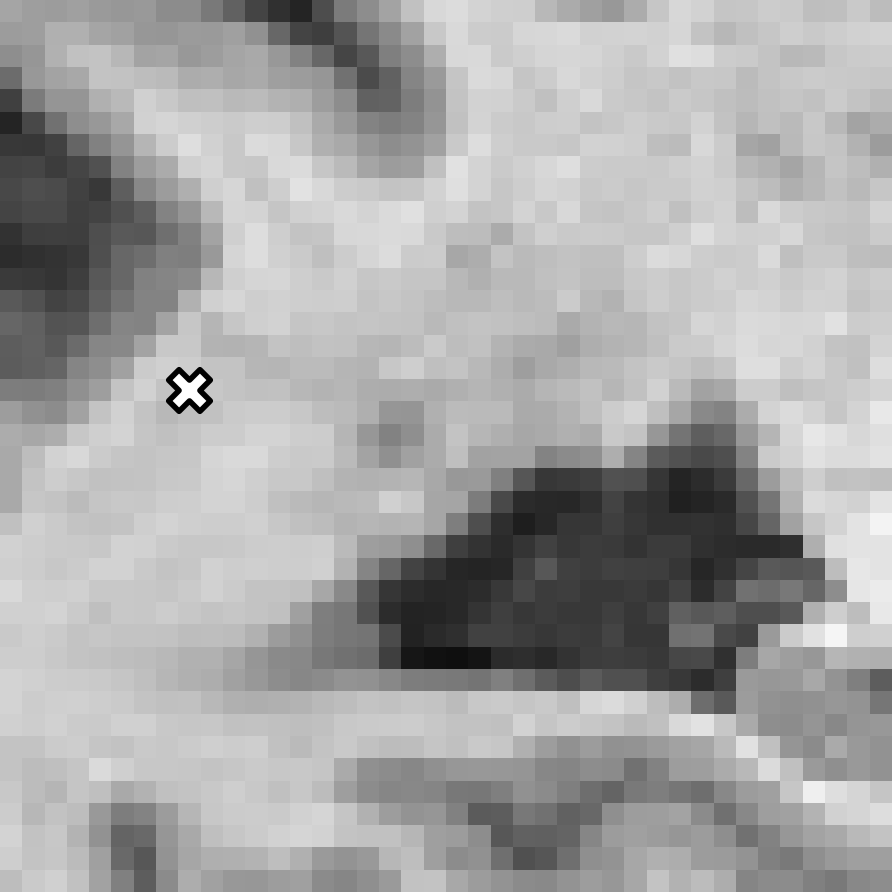} &
    \includegraphics[height=\myheight]{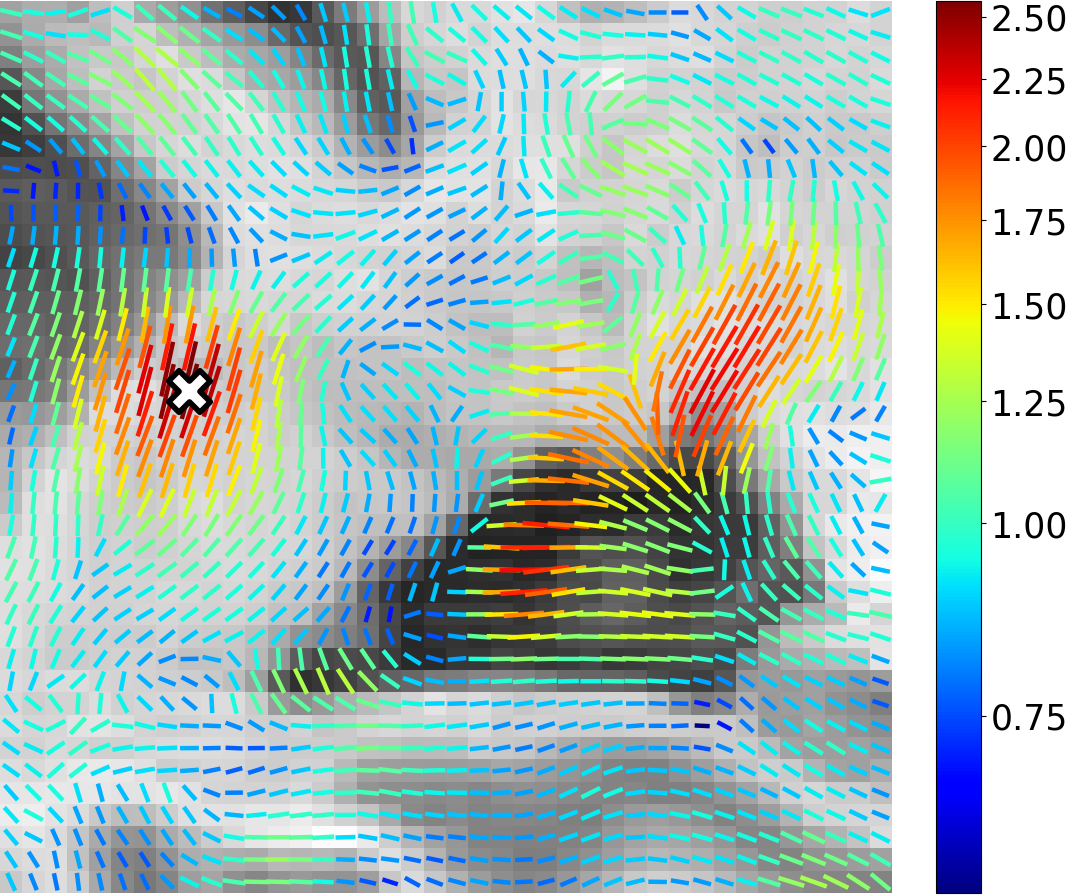} \\   
    \includegraphics[height=\myheight]{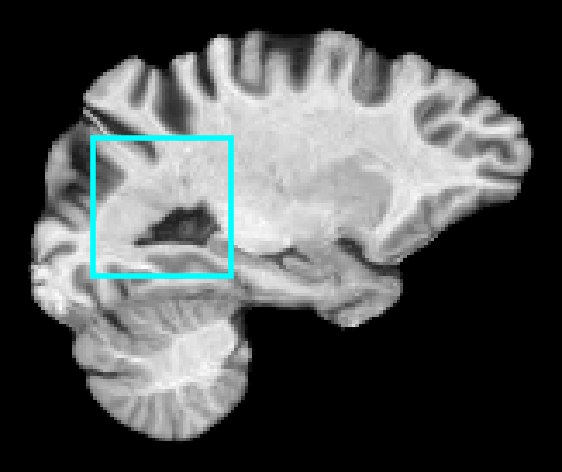} &
    \includegraphics[height=\myheight]{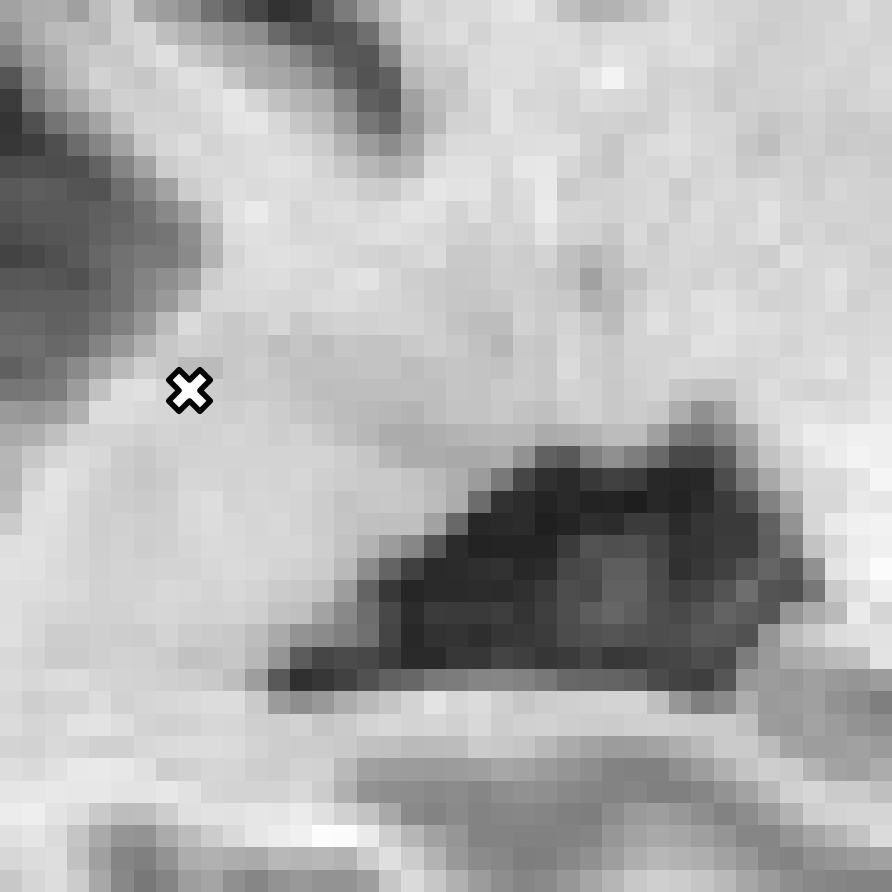} &          
    \includegraphics[height=\myheight]{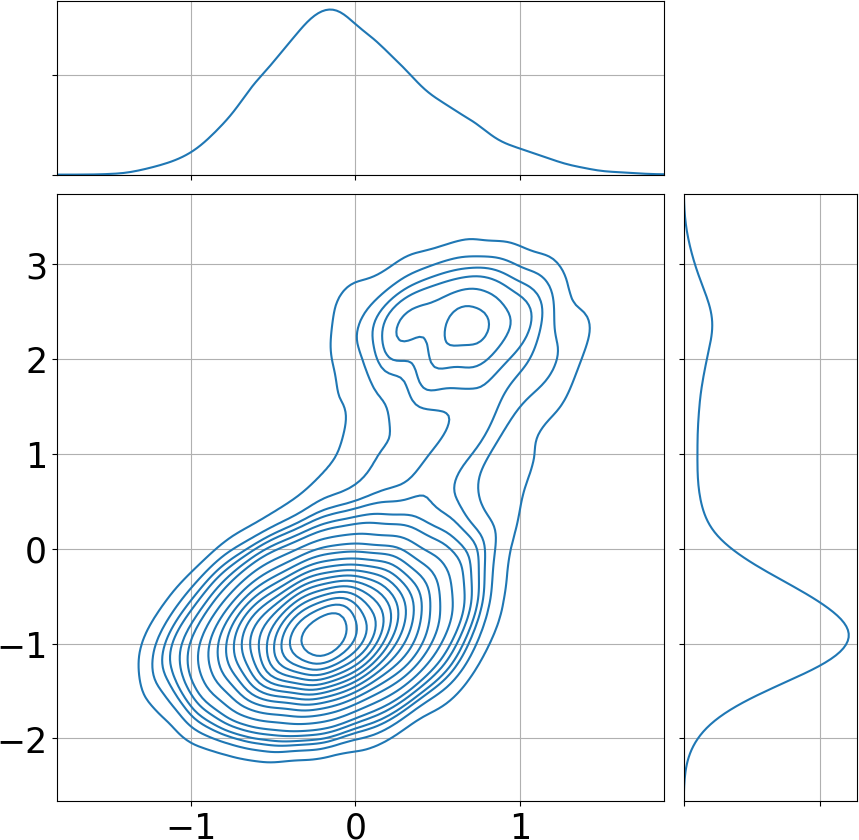}
  \end{tabular}
  \caption{Same as Fig.~\ref{fig:case4}, but for a monomodal registration problem (T1-T1) in a different pair of subjects.}
  \label{fig:case2}
\end{figure*}



%
%
Since sampling is a slow procedure,
we skull-stripped the images 
and cropped them 
to focus only on the brain,
reducing the number of voxels by approximately a factor of four
compared to the original image size.
For each registration pair,
we 
simulated a single MCMC chain, initialized at the deformation produced by the optimizer 
without 
applying
the 
Jacobian-constraining
projection step.
%
The regularization hyperparameter $\gamma$ was set to
its
task-specific
tuned value, 
i.e., $1 \cdot 10^{-6}$ for T1-T2 and $2.5 \cdot 10^{-7}$ for T1-T1 (see Table~IV of the supplementary material).


%
Although 
the draws produced by the sampler are full 3D deformation fields,
we visualize the results only for a single 2D slice selected for each case,
and analyze 
the two components of the deformation corresponding to the visible directions in the selected slice.
%
%
%
%
%
To establish whether the chain had approximately converged to a stationary distribution after running it for a number of iterations, we 
discarded the first half of the chain as a warm-up period.
The remaining draws were then used to compute,
for each 
brain
voxel 
in the 
selected slice,
the split-$\hat{R}$ statistic~\cite{Gelman2020_chapter11} for 
each directional 
component of the 
deformation 
vector.
%
This procedure splits the remaining chain again in two halves,
and checks whether
the variances of the draws within each half
are similar to the variance of the 
pooled draws.
When this is the case split-$\hat{R}$ will approach 1, with a value of $< 1.1$ often taken as evidence of convergence 
in practice~\cite{Gelman2020_chapter11}.
%
%
%
This was indeed the case for the first three cases (Figs.~\ref{fig:case4} and ~\ref{fig:case2}, and Fig.~8 of the supplementary material), in which split-$\hat{R}$ values in both directional components were lower than 1.03 across all voxels after 
around 24 million iterations, which took approximately 12 hours of computation time (see Sec.~X of the supplementary material for visualizations of the split-$\hat{R}$ maps that we obtained).
For the fourth case (Fig.~9 of the supplementary material), however, the chain was still non-stationary -- with split-$\hat{R}$ values over 1.6 in some areas -- even after 45 million iterations (around 23 hours), after which the 
sampling procedure
was stopped.


We note that
more
formally
checking 
convergence with split-$\hat{R}$
would 
ideally
involve
not only
assessing stationarity 
in 
3D
-- rather than in just a single 2D slice as we have done --
but also initializing multiple chains from different starting points 
to verify
that they 
converge to the same 
distribution~\cite{Gelman2020_chapter11}.
Nevertheless,
we believe 
the results shown 
in Figs.~\ref{fig:case4} and ~\ref{fig:case2} and Figs.~8 and~9 of the supplementary material
already 
provide an indication of the difficulty of characterizing the posterior distribution 
in nonlinear brain registration:
For each case, we identified the point with the highest variability 
across the samples
within a selected $40\times40$ ROI 
(the white cross in the plots), and visualize the 2D marginal distribution at that point. Across all cases 
-- 
including in the T1-T1 inter-subject registration that 
has been the typical focus of
the probabilistic registration literature 
-- 
this distribution is clearly not Gaussian and often strongly multimodal (multiple peaks). This raises doubts regarding the appropriateness of the frequently-used variational approach to quantify registration uncertainty in such scenarios.

\fi 

\section{Discussion and Conclusion}

In this paper, we proposed a new probabilistic model for multimodal registration that 
facilitates
both 
optimization and MCMC sampling 
through the use of latent variables.
When applied to nonlinear registration, 
we showed experimentally that 
this enables
robust 
registration performance
across very different 
scenarios
with minimal user tuning.
We also demonstrated 
effective sampling 
from 
complex, 
very high-dimensional
deformation posteriors
in both monomodal and multimodal registration settings alike
-- to the best of our knowledge 
for the first time
in the literature.


%
Although we demonstrated the proposed techniques on 
DCT basis functions
that are
regularized 
using
second-order derivatives,
the 
same
mechanism translates directly to settings with different parameterizations and regularizations, 
including 
those involving
affine transformations
(when linear basis functions are used)
or 
those
where
first-order 
derivatives 
are penalized
(so-called membrane energy).
Further generalizing 
the method
to cases in which 
different
directional components
of the transformation model
need to be updated jointly,
such as rigid transformations or models in which linear-elastic energy
is penalized,
should also be straightforward.

In the optimization variant of our method,
we ensured invertibility of the 
computed
deformation fields 
by projecting them onto the nearest fields 
with bounded
Jacobian determinants.
%
A more elegant solution that would also likely produce smoother warps~\cite{Lange2020} 
would 
use diffeomorphic parameterizations instead.
Given the close resemblance between the 
deformation updates in the
demons algorithm and 
the ones in 
ours,
it may be possible to apply the same mechanism as those used to make the demons algorithm diffeomorphic~\cite{Vercauteren2009} to our setting.

During the course of our experiments, we encountered 
some
difficulties that have not received much attention in the probabilistic image registration literature. 
For instance,
it is 
fairly straightforward
to 
sample also 
from
the regularization strength $\gamma$ 
in the proposed sampler
--
in the simplest case\footnote{We also tested a collapsed Gibbs sampler that integrates out $\fat{c}_d$ analytically to sample from $p(\gamma| \fat{n}) \propto p(\fat{n}| \gamma) p(\gamma)$ instead.
}, by sampling from
$p(\gamma | \{\fat{c}_d\} ) \propto \prod_d p(\fat{c}_d| \gamma) p(\gamma)$
for some prior $p(\gamma)$, since $\prod_d p(\fat{c}_d| \gamma)$ has the form of a Gamma distribution over $\gamma$. 
%
Automatically inferring
the regularization strength
has been successfully performed in the literature~\cite{%
RisholmWBIR2010,
RisholmMEDIA2013,
SimpsonNI2012,
SimpsonMEDIA2015,
LeFolgocMEDIA2017,
LeFolgocTMI2017,
Grzech2021},
and would remove the last remaining user-tuned hyperparameter from our method.
%
Unfortunately, 
in practice we observed that this only worked robustly in a subset of the registration scenarios that we considered: for brain scans involving 
non-T1-weighted contrasts, 
in particular,
the sampler was often drawn to
very low $\gamma$ values,
resulting in 
significant drops in registration accuracy. 
We note that automatically
inferring regularization strengths
has only been attempted 
in low-dimensional deformation models 
in prior work 
--
typically involving no more than a few thousand parameters, 
which makes the deformations well-regularized
even for very low regularization strengths.
We speculate that this success may not 
extend to dense 3D deformation fields,
where the model has the option to select millions of 
effective
degrees of freedom instead.


Another unresolved issue pertains to a virtual decimation factor that has been used by some authors in the probabilistic registration literature~\cite{%
SimpsonNI2012,
SimpsonMEDIA2015,
LeFolgocMEDIA2017,
LeFolgocTMI2017,
Grzech2021}.
It aims to compensate misspecifications of the forward model -- namely the assumption that the voxel intensities in the fixed image are conditionally independent given the moving image -- by 
down-weighting the log-likelihood function in
the 
model.
However, this technique 
rests on 
strong assumptions itself (in particular, 
that the registration is monomodal, with 
spatial correlations 
arising
from blurring with a Gaussian kernel~\cite{Worsley1995,Groves2011}),
and 
it is 
currently
unclear how it 
could be generalized 
to
multimodal 
settings.


Although we have demonstrated the feasibility of sampling from dense 3D deformation fields in this paper, future work should address two remaining shortcomings of the proposed 
sampler. 
First, a key step in our derivations is the approximation of a cubic B-spline with a Gaussian distribution. 
While this approximation 
appears to be 
sufficiently accurate for obtaining a robust and efficient EM optimizer, its effect on the corresponding MCMC sampler is harder to evaluate. One way to obtain an exact sampling method 
is to replace the current spatial interpolation model, which is based on discrete latent variables distributed according to a B-spline kernel, with a model based on continuous latent variables that are Gaussian distributed: node assignments would then be based on whether the latent variable falls within the cuboid surrounding each node. 
This, however, 
requires an additional step to sample from a truncated Gaussian during inference, and was not attempted for this paper.

A second 
remaining
issue is that the proposed sampler operates only at full image resolution, which can lead to slow mixing and limit exploration to regions near a few posterior modes. In contrast, 
for optimization we 
construct 
a coarse-to-fine hierarchy
in which both the fixed and moving images are repeatedly downsampled,
as is commonly done in the field~\cite{thevenaz1998pyramid}.
Downsampling the fixed image dramatically reduces the computational cost, 
since the number of 
voxels in the model
decreases by a factor of 8 with each downsampling step in 3D.
Downsampling the moving image, on the other hand, 
broadens the effective B-spline support in our approach,
thereby
smoothing the optimization landscape and helping the optimizer escape local optima.

Several approaches exist 
in the MCMC literature
that use related hierarchical strategies to improve exploration and/or lower computational cost. 
Tempering methods, for example, use progressively flatter versions of the target distribution to make the transitions between modes easier \cite{marinari1992simulated,geyer1991markov}. Multilevel delayed-acceptance MCMC instead uses a recursive hierarchy of increasingly accurate models, where proposals at each finer level are generated using subchains from the cheaper, coarser level and then corrected to preserve the fine-level target distribution \cite{lykkegaard2023multilevel}. Multilevel MCMC uses a hierarchy of resolutions to construct a telescoping estimator, allowing more samples to be drawn from inexpensive coarse levels and fewer from expensive fine levels \cite{dodwell2015hierarchical}. 
%
%
%
Designing a hierarchical strategy that draws on these ideas 
to make 
our sampler more efficient
is an open area for future research.
%
%
%


\noteToSelf{Possible structure of discussion:
\begin{itemize}
    \item What we proposed: Say that 1) we have a probabilistic model for MI registration with a dedicated optimizer (EM) and 2) a dedicated MCMC sampler.
    \item Results better or on par to benchmark method on a variety of datasets. 2) only one parameter to tune (gamma, or even zero!)
    \item Limitations:
    \begin{itemize}
        \item (Say something that you can get better performance by training CNNs on specific tasks, but again they have difficulty generalizing to unseen data)
        \item Comparison limited by grid-search. Maybe better parameters for benchmark methods.
        \item No symmetric and diffeomorphic field by design. [So not good if focus is studyign deformation fields (e.g., computational neuroanatomy) but OK for us since focus is on \emph{multimodal} registration ]
        \item Sampler slow and might not explore the true posterior distribution.
    \end{itemize}
    \item Extensions/Future work:
    \begin{itemize}
        \item Although not tested here, affine and rigid transformations can be computed (code is already available). Same for flexible similarity metrics
        \item Faster and less stuck sampler: Multi-resolution sampler
    \end{itemize}
\end{itemize}
}

\noteToSelf{
In this paper, we have proposed a novel mutual information nonlinear registration method with a dedicated EM optimization scheme. Furthermore, we derived a dedicated MCMC sampler, allowing us to report uncertainty estimates of the model's parameters. 
}

\noteToSelf{
We validated the method on several datasets of brain, abdomen, and lung images. Our results indicate that the method has comparable or even better performance than benchmark methods while using the same set of hyperparameters. 
}

\noteToSelf{
The work presented here has several limitations. First, our comparison of the proposed method and benchmark methods is limited to the number of hyperparameters we used in our grid-search analysis. A more extensive comparison is needed to confirm our findings. 
Second, the proposed method is not symmetric nor diffeomorphic by design. For diffeomorphic: say something like FNIRT wiki, they made it sound like an advantage. For symmetry, future work.
Finally, the uncertainty maps produced by the method are 1) slow and 2) might not completely explore the full posterior distribution, due to the high dimensionality of the parameter space. Future work is needed to design samplers that can leverage multiple resolution levels, as in our optimization scheme. 
}

\noteToSelf{Say something about EndtoEnd version used only for brain images. Results should be confirmed on other organs, and current implementation (center of mass) can be improved for more robust results.}

\noteToSelf{Results from Learn2Reg:

Abdomen CT-CT (different dataset, labels?, both inter): (Note that Hauss95 in Learn2Reg doesn't count for voxel resolutions != 1 (bug???))\\
ConvexAdam (best): Dice: 0.69, Hauss95: 11.03\\
CorrField: Dice: 0.49, Hauss95: 17.22\\
Abdomen MR-CT (intra!):\\
ConvexAdam (best): Dice: 0.75, Hauss95: 24.92\\
CorrField: Dice: 0.76, Hauss95: 23.35\\
Brain (T1-T1) (skull stripped! Different labels?):\\ 
ConvexAdam (best): Dice: 0.81, Hauss95: 1.63\\
CorrField: Dice: 0.74, Hauss95: 2.36\\
}


\section*{Acknowledgment}

Stefano Cerri: Lundbeck Foundation (R449-2023-1512). 
Ya\"el Balbastre: Royal Society (NIF\textbackslash{}R1\textbackslash{}232460).

\bibliographystyle{IEEEtran}
\bibliography{ref}

\appendices


\if\hideResults0

\fi 

\includepdf[pages=-]{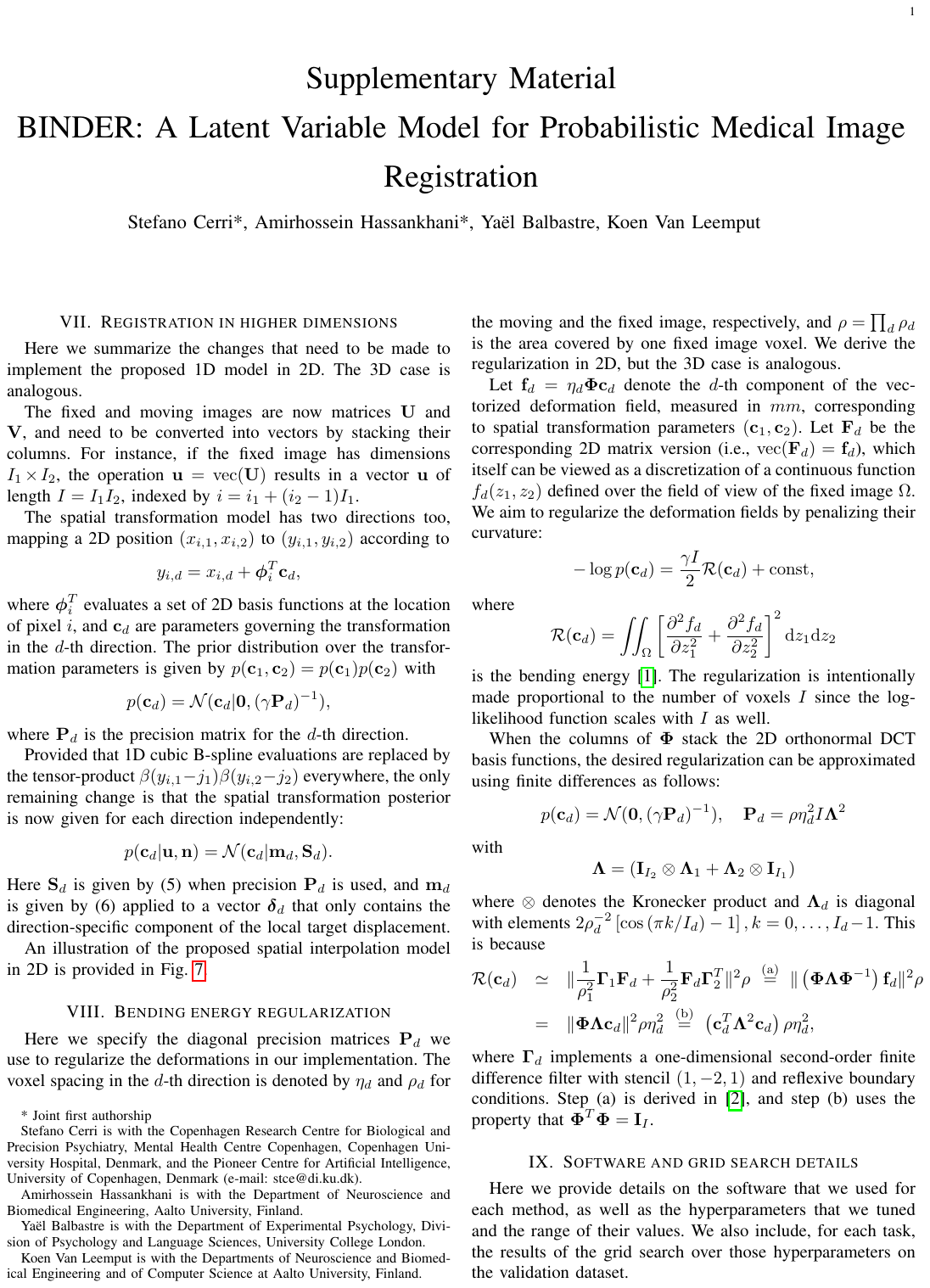}

\end{document}